%% file: main.tex
\newif\ifdouble
\newif\ifsingle
\newif\ifchange
\doubletrue

\ifdouble
  \documentclass[sigconf, usenames, dvipsnames]{acmart}
\else
  \documentclass[manuscript,anonymous]{acmart}
  \singletrue
\fi

\usepackage[utf8x]{inputenc}
\usepackage{fontawesome5}

\usepackage{dblfloatfix} 
\usepackage{booktabs} 
\usepackage{arydshln}
\usepackage{subfig}
\ifchange
\usepackage{soul}

\else

\fi

\newcommand{\sub}[1]{\vspace{0.2cm} \noindent \textbf{\textit{#1: }}}
\newcommand{\subsub}[1]{\vspace{0.1cm} \noindent \textit{--- #1}: }

\newif\ifTopLevelColor
\newif\ifCategoryColor
\newif\ifFullColor
\newif\ifSubSubColor

\CategoryColortrue
\FullColortrue
\SubSubColortrue

\input{color-coding}

\input{figures}

\AtBeginDocument{%
  \providecommand\BibTeX{{%
    \normalfont B\kern-0.5em{\scshape i\kern-0.25em b}\kern-0.8em\TeX}}}

\copyrightyear{2022}
\acmYear{2022}
\setcopyright{acmlicensed}\acmConference[CHI '22]{CHI Conference on Human Factors in Computing Systems}{April 29-May 5, 2022}{New Orleans, LA, USA}
\acmBooktitle{CHI Conference on Human Factors in Computing Systems (CHI '22), April 29-May 5, 2022, New Orleans, LA, USA}
\acmPrice{15.00}
\acmDOI{10.1145/3491102.3517719}
\acmISBN{978-1-4503-9157-3/22/04}

\begin{document}
\pagenumbering{arabic}
\pagestyle{plain}
\title{Augmented Reality and Robotics: A Survey and Taxonomy for AR-enhanced Human-Robot Interaction and Robotic Interfaces}

\author{Ryo Suzuki}
\affiliation{%
  \institution{University of Calgary}
  \streetaddress{Address}
  \city{Calgary, AB}
  \country{Canada}}
\email{ryo.suzuki@ucalgary.ca}

\author{Adnan Karim}
\affiliation{%
  \institution{University of Calgary}
  \streetaddress{Address}
  \city{Calgary, AB}
  \country{Canada}}
\email{adnan.karim@ucalgary.ca}

\author{Tian Xia}
\affiliation{%
  \institution{University of Calgary}
  \streetaddress{Address}
  \city{Calgary, AB}
  \country{Canada}}
\email{tian.xia2@ucalgary.ca}

\author{Hooman Hedayati}
\affiliation{%
  \institution{UNC at Chapel Hill}
  \streetaddress{Address}
  \city{Chapel Hill, NC}
  \country{U.S.A.}}
\email{hooman@cs.unc.edu}

\author{Nicolai Marquardt}
\affiliation{%
  \institution{University College London}
  \streetaddress{Address}
  \city{London}
  \country{U.K.}}
\email{n.marquardt@ucl.ac.uk}

\renewcommand{\shortauthors}{Suzuki, et al.}

\input{0-abstract}

\FigureTeaser

\maketitle

\input{1-introduction}
\input{2-scope}
\input{3-approach}

\input{4-robots}

\input{5-purpose}

\input{6-information}

\input{7-design}
\input{8-interactions}

\input{9-applications}

\input{10-evaluation}
\input{11-discussion}
\input{12-opportunity}
\input{13-conclusion}




\ifdouble
  \balance
\fi
\bibliographystyle{ACM-Reference-Format}
\bibliography{references}

\clearpage
\newcommand{\TableFontSize}{\footnotesize}
\newcommand{\TableConfig}{p{0.15\textwidth} c p{0.68\textwidth}}
\newcommand{\TableHeader}{Category & Count & Citations}
\input{14-appendix}

\clearpage

\end{document}

%% file: color-coding.tex
\definecolor{color1}{HTML}{3d3ca4} 
\definecolor{color2}{HTML}{308377} 
\definecolor{color3}{HTML}{3c88a4} 
\definecolor{color4}{HTML}{9a6003} 
\definecolor{color5}{HTML}{359266} 
\definecolor{color6}{HTML}{359091} 
\definecolor{color7}{HTML}{84005f} 
\definecolor{color8}{HTML}{926735} 
\definecolor{color9}{HTML}{1b1b1b} 

\definecolor{color10}{HTML}{16697A} 

\newcommand{\subsubmark}[1]{$\textcolor{#1}{\bullet}$}

\ifTopLevelColor
\newcommand{\coloredSection}[2]{\textcolor{#1}{\section{#2}}\noindent\kern-0.1ex}
\else
\newcommand{\coloredSection}[2]{\section{#2}}
\fi

\newcommand{\approach}[3]{\sub{\submark{#3} Approach-#1. #2}}
\newcommand{\robot}[2]{\sub{\submark{color3} Dimension-#1. #2}}
\newcommand{\purpose}[2]{\sub{\submark{color4} Purpose-#1. #2}}
\newcommand{\information}[2]{\sub{\submark{color5} Information-#1. #2}}
\newcommand{\design}[2]{\sub{\submark{color6} Design-#1. #2}}
\newcommand{\interaction}[3]{\sub{\submark{color7} Dimension-#1. #2}}

\newcommand{\opportunity}[2]{\sub{\submark{color10} Opportunity-#1. #2}}

\ifCategoryColor
\renewcommand{\approach}[3]{\sub{\textcolor{#3}{Approach-#1.} #2}}
\renewcommand{\robot}[2]{\sub{\textcolor{color3}{Dimension-#1.} #2}}
\renewcommand{\purpose}[2]{\sub{\textcolor{color4}{Purpose-#1.} #2}}
\renewcommand{\information}[2]{\sub{\textcolor{color5}{Information-#1.} #2}}
\renewcommand{\design}[2]{\sub{\textcolor{color6}{Design-#1.} #2}}
\renewcommand{\interaction}[3]{\sub{\textcolor{color7}{Dimension-#1.} #2}}

\renewcommand{\opportunity}[2]{\sub{\textcolor{color10}{Opportunity-#1.} #2}}
\fi

\ifFullColor
\renewcommand{\approach}[3]{\sub{\textcolor{#3}{Approach-#1. #2}}}
\renewcommand{\robot}[2]{\sub{\textcolor{color3}{Dimension-#1. #2}}}
\renewcommand{\purpose}[2]{\sub{\textcolor{color4}{Purpose-#1. #2}}}
\renewcommand{\information}[2]{\sub{\textcolor{color5}{Information-#1. #2}}}
\renewcommand{\design}[2]{\sub{\textcolor{color6}{Design-#1. #2}}}
\renewcommand{\interaction}[3]{\sub{\textcolor{color7}{Dimension-#1. #2}}}

\renewcommand{\opportunity}[2]{\sub{\textcolor{color10}{Opportunity-#1. #2}}}
\fi

\ifSubSubColor
\newcommand{\approachsub}[2]{\subsub{\textcolor{#2}{#1}}}
\newcommand{\designsub}[1]{\subsub{\textcolor{color6}{#1}}}
\newcommand{\interactionsub}[1]{\subsub{\textcolor{color7}{#1}}}

\newcommand{\opportunitysub}[1]{\subsub{\textcolor{color10}{#1}}}
\else
\renewcommand{\subsub}[1]{\vspace{0.1cm} \noindent \textit{#1}: }
\newcommand{\approachsub}[2]{\subsub{\subsubmark{#2} #1}}
\newcommand{\designsub}[1]{\subsub{\subsubmark{color6} #1}}
\newcommand{\interactionsub}[1]{\subsub{\subsubmark{color7} #1}}
\fi

%% file: figures.tex
\ifdouble
  
  \newcommand{\FigureHalfWidth}{\columnwidth}
\else 
  
  \newcommand{\FigureHalfWidth}{0.7\textwidth}
\fi

\newcommand{\FigureTeaser}{
\begin{teaserfigure}
\ifdouble
\fi
\includegraphics[width=\textwidth]{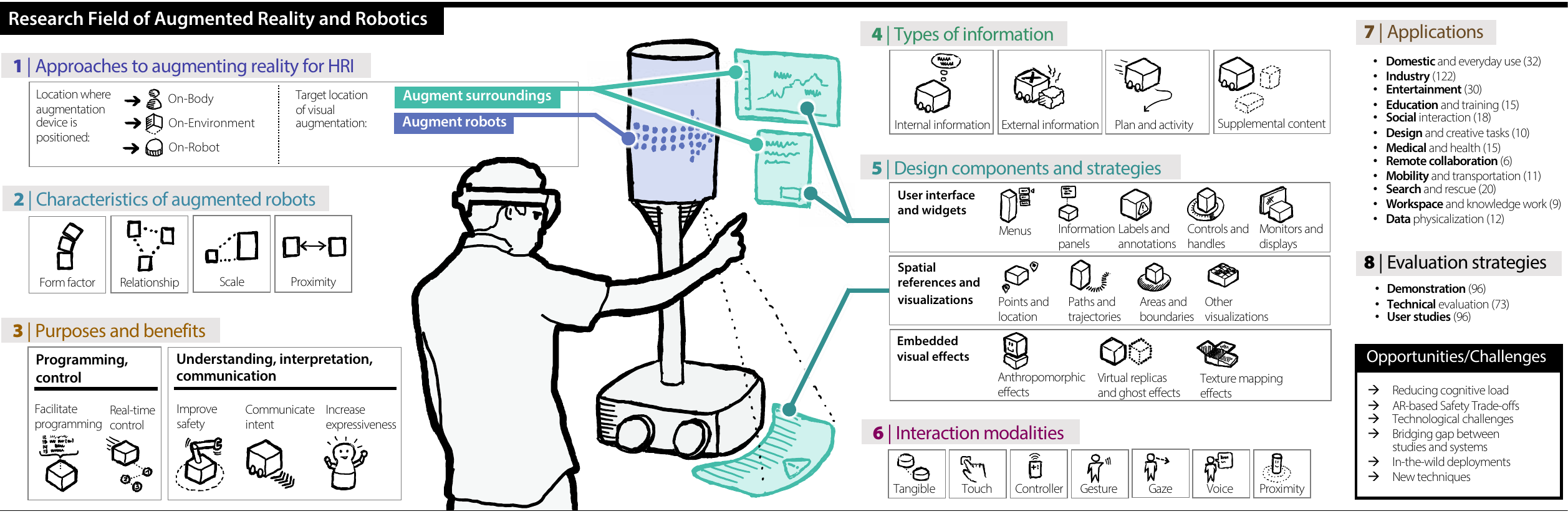}
\caption{Visual abstract of our survey and taxonomy of augmented reality interfaces used with robotics, summarizing eight key dimensions of the design space. 
All sketches and illustrations are made by the authors (Nicolai Marquardt for Figure~\ref{fig:teaser} and~\ref{fig:approaches} and Ryo Suzuki  for Figure~\ref{fig:approaches}-\ref{fig:future-opportunity-4}) and are available under CC-BY 4.0\ \faCreativeCommons\ \faCreativeCommonsBy\ with the credit of original citation.
All materials and an interactive gallery of all cited papers are available at \url{https://ilab.ucalgary.ca/ar-and-robotics/}
}
\Description[]{}
\label{fig:teaser}
\end{teaserfigure}
}

\newcommand{\FigureScreenShots}{
\begin{figure*}[h!]
\centering
\includegraphics[width=1\textwidth]{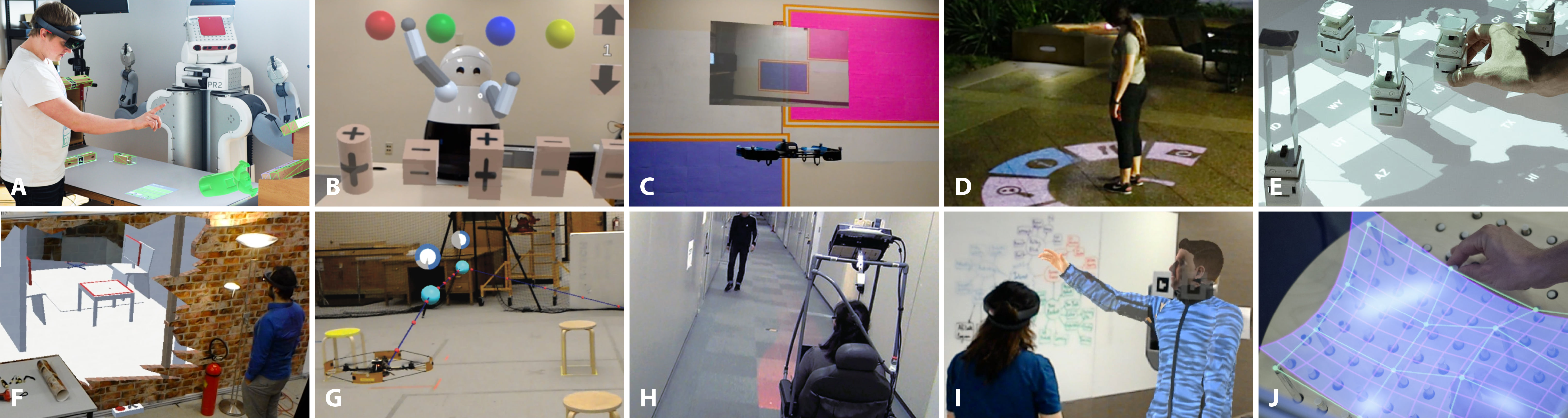}
\caption{Examples of augmented reality and robotics research: 
A) collaborative programming~\cite{bambusek2019combining}, 
B) augmented arms for social robots~\cite{groechel2019using},
C) drone teleoperation~\cite{hedayati2018improving},
D) a drone-mounted projector~\cite{cauchard2019drone},
E) projected background for data physicalization~\cite{suzuki2019shapebots}, 
F) drone navigation for invisible areas~\cite{erat2018drone},
G) trajectory visualization~\cite{walker2018communicating}, 
H) motion intent for pedestrian~\cite{watanabe2015communicating}, 
I) a holographic avatar for telepresence robots~\cite{jones2021belonging}, 
J) surface augmentation for shape displays~\cite{leithinger2013sublimate}.
}
\label{fig:screenshots}
\end{figure*}
}

\newcommand{\FigureApproach}{
\begin{figure*}[h!]
\centering
\resizebox{\textwidth}{!}{
\includegraphics[height=\textwidth]{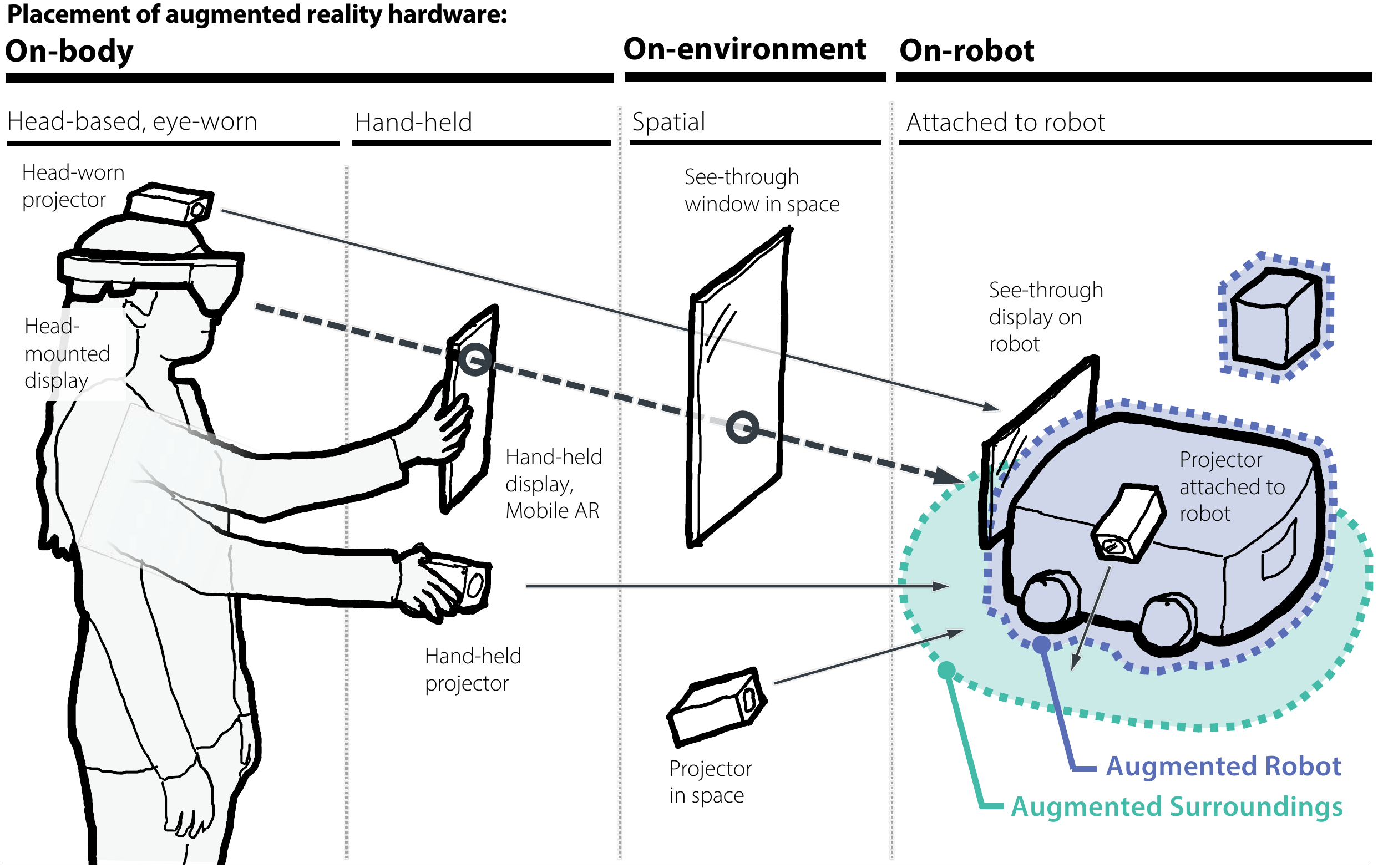}
\hspace{0.8cm}
\includegraphics[height=\textwidth]{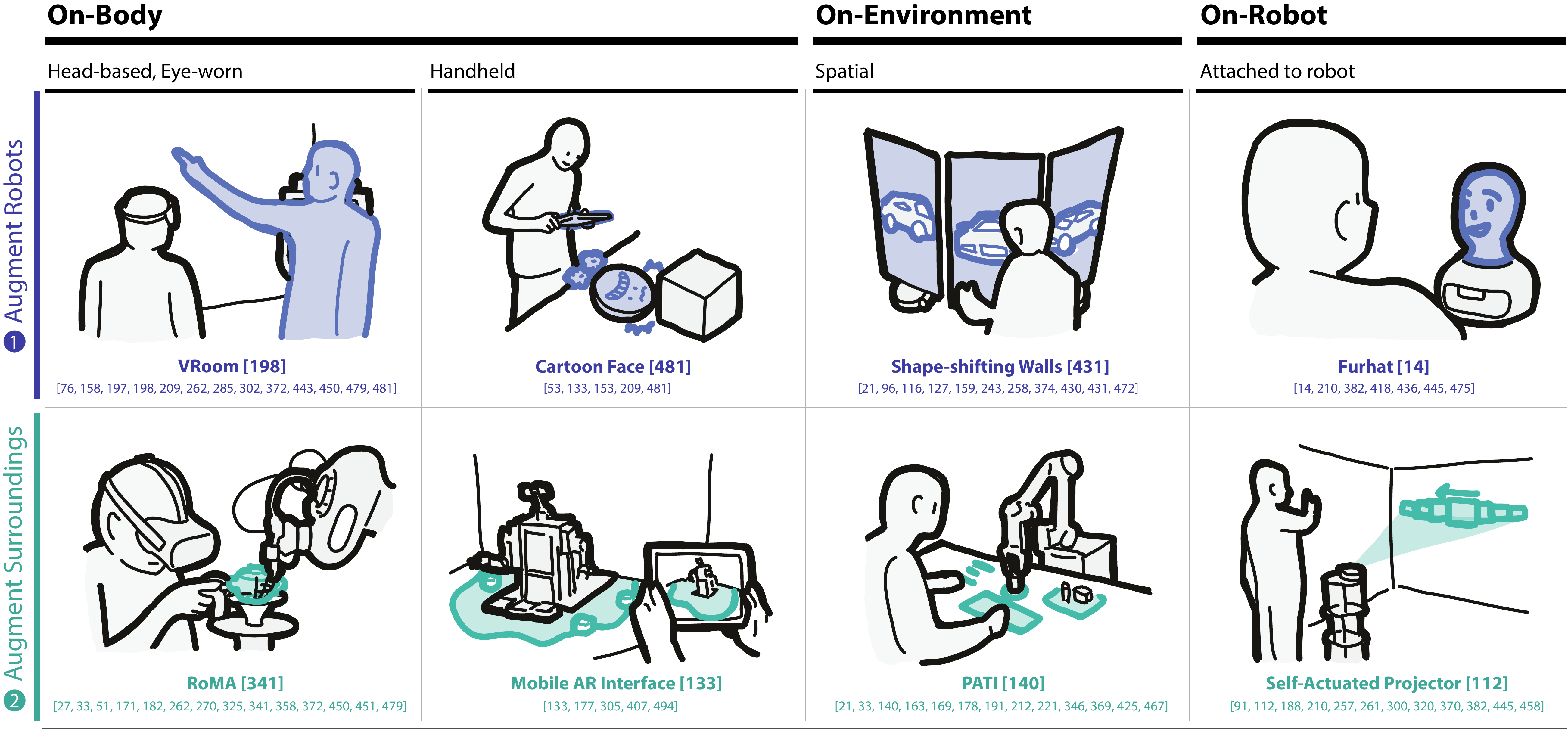}
}
\caption{Approaches to augmenting reality in robotics.}
~\label{fig:approaches}
\end{figure*}
}

\newcommand{\FigureRobot}{
\begin{figure}[h!]
\centering
\resizebox{\FigureHalfWidth}{!}{
\includegraphics[height=\textwidth]{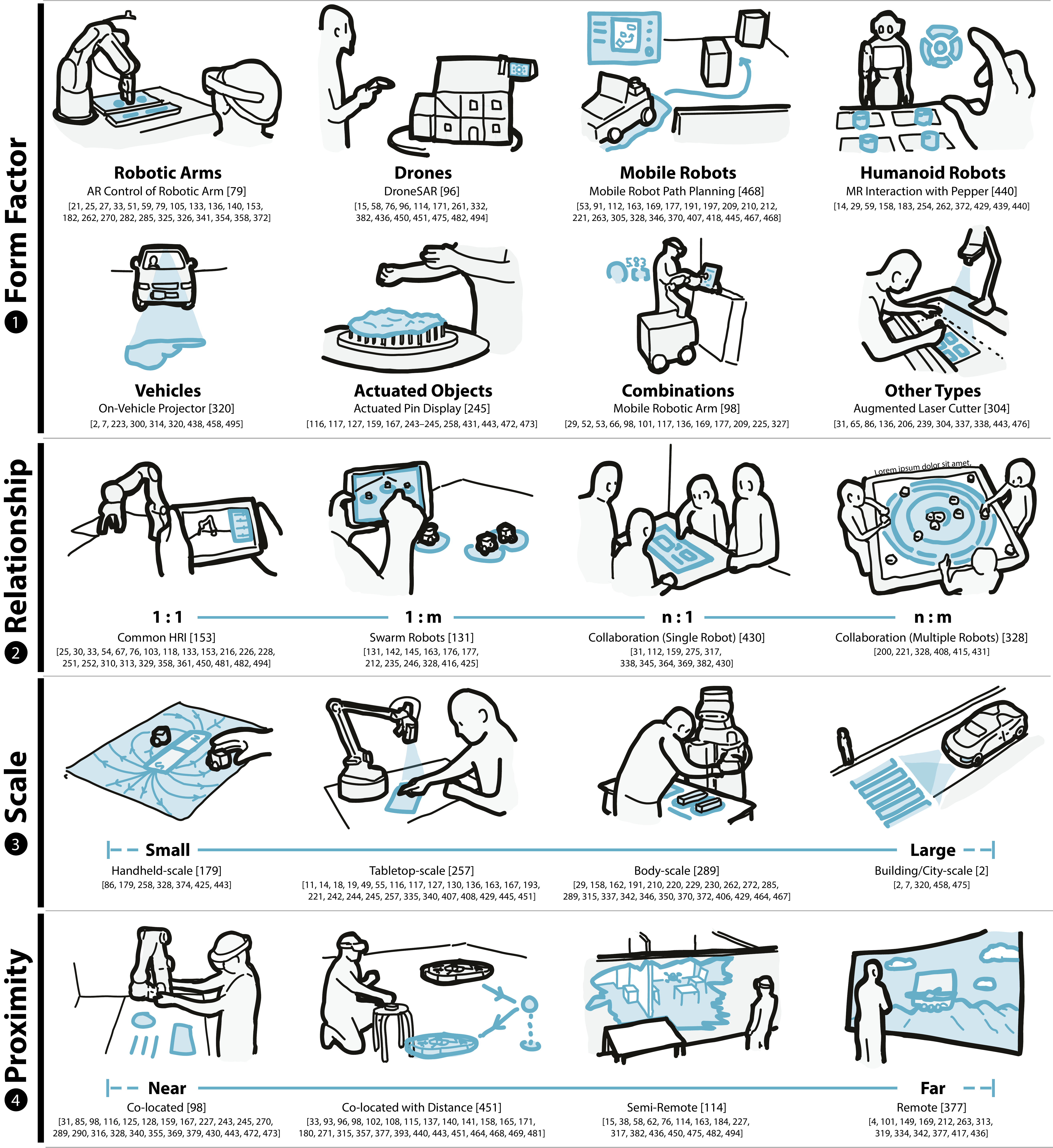}
}
\caption{Characteristics of augmented robots.}
~\label{fig:robots}
\end{figure}
}

\newcommand{\FigurePurpose}{
\begin{figure*}[!h]
\centering
\includegraphics[width=\textwidth]{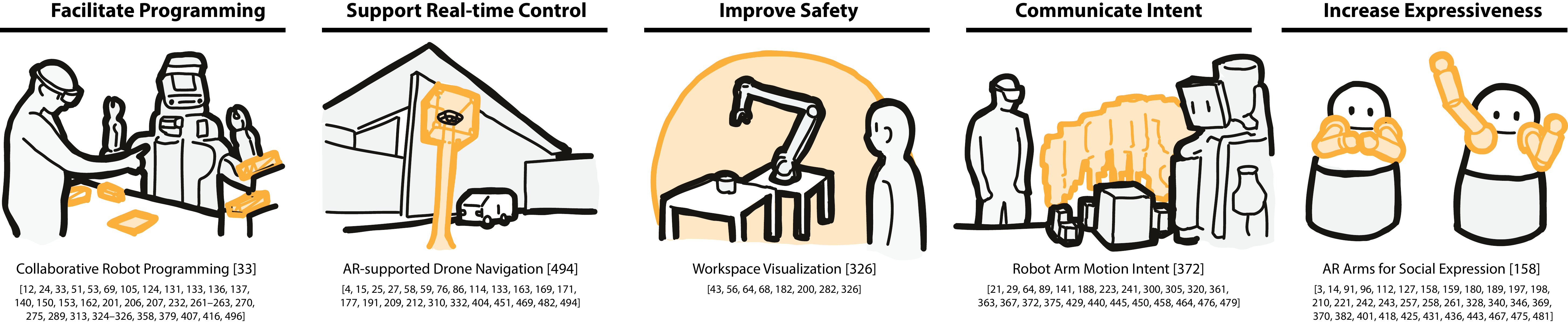}
\caption{Purposes and benefits of visual augmentation.}
~\label{fig:purposes}
\end{figure*}
}

\newcommand{\FigureInformation}{
\begin{figure*}[!b]
\centering
\resizebox{\textwidth}{!}{
\includegraphics[height=\textwidth]{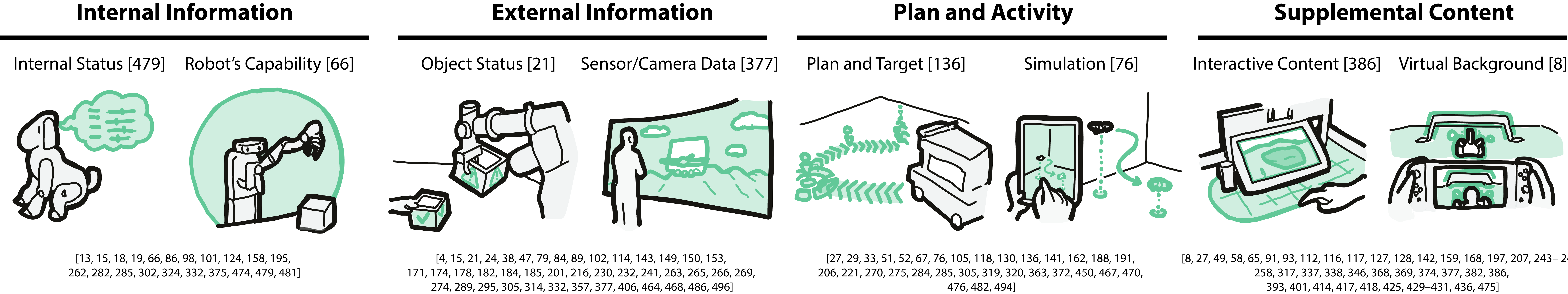}
}
\caption{Types of presented information}
~\label{fig:information}
\end{figure*}
}

\newcommand{\FigureDesign}{
\begin{figure*}[!h]
\centering
\resizebox{\textwidth}{!}{
\includegraphics[height=\textwidth]{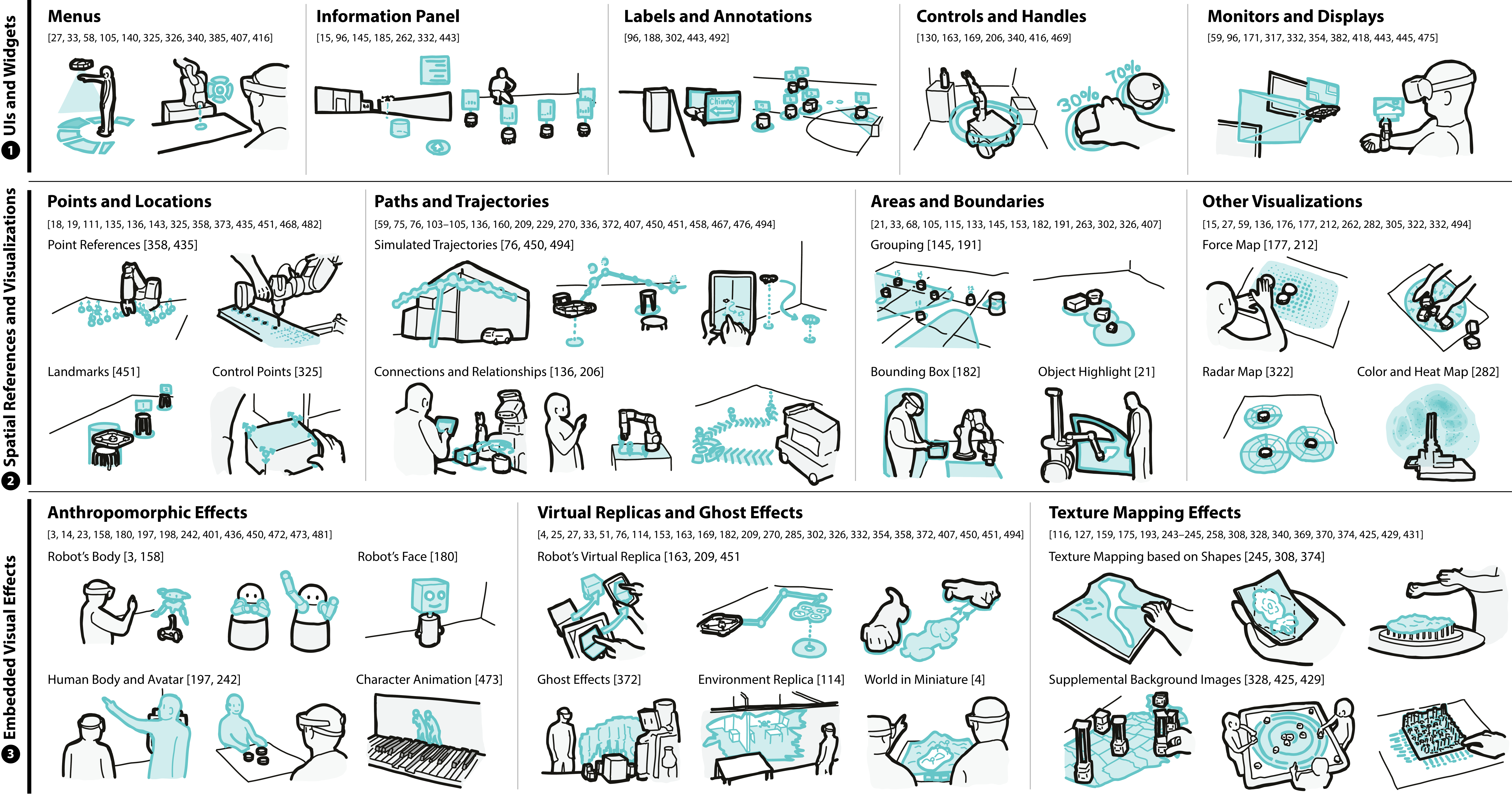}
}
\caption{Design components and strategies for visual augmentation}
~\label{fig:design}
\end{figure*}
}

\newcommand{\FigureInteractivity}{
\begin{figure}[h!]
\centering
\includegraphics[width=\FigureHalfWidth]{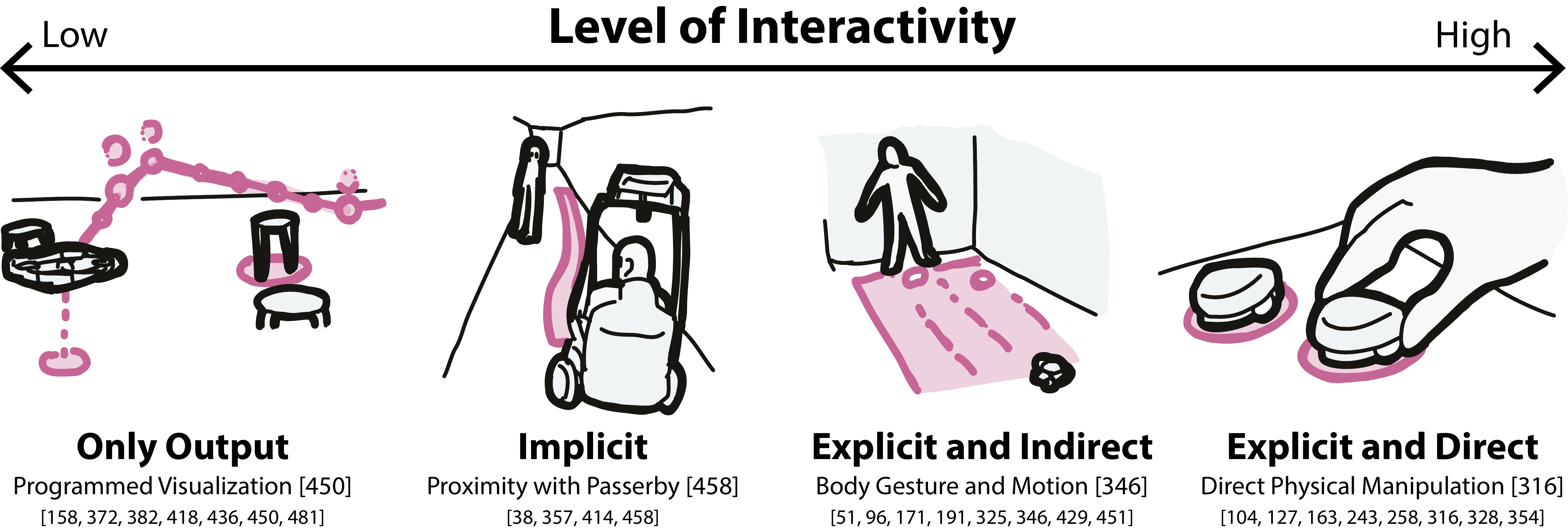}
\caption{Level of interactivity}
~\label{fig:interactivity}
\end{figure}
}

\newcommand{\FigureInteraction}{
\begin{figure*}[!h]
\centering
\includegraphics[width=\textwidth]{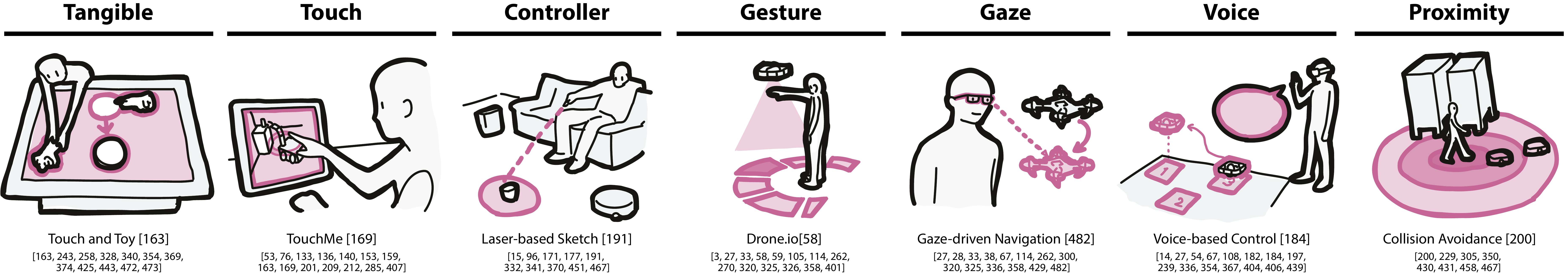}
\caption{Interaction modalities and techniques}
~\label{fig:interactions}
\end{figure*}
}

\newcommand{\FigureApplication}{
\begin{figure*}[!h]
\centering
\resizebox{\textwidth}{!}{
\includegraphics[height=\textwidth]{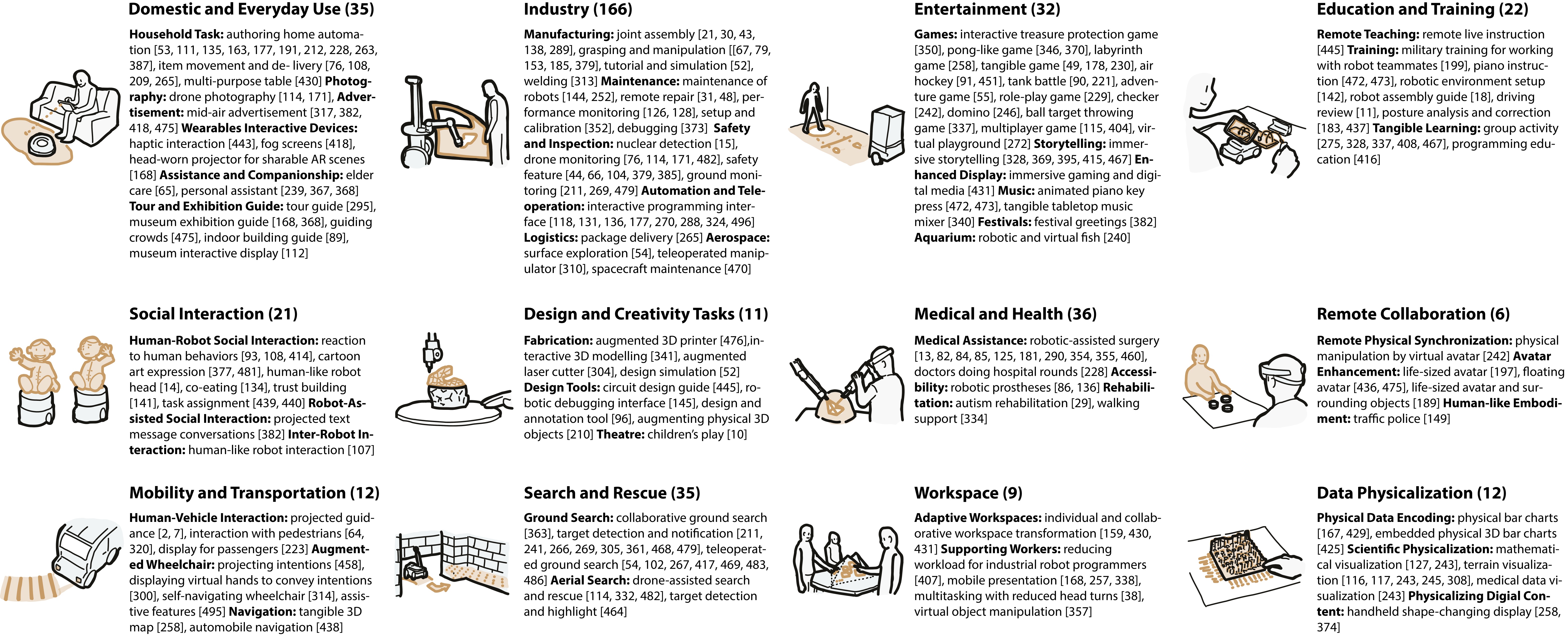}
}
\caption{Use cases and application domains}
~\label{fig:applications}
\end{figure*}
}

\newcommand{\FigureOpportunityOne}{
\begin{figure}[!h]
\centering
\includegraphics[width=\FigureHalfWidth]{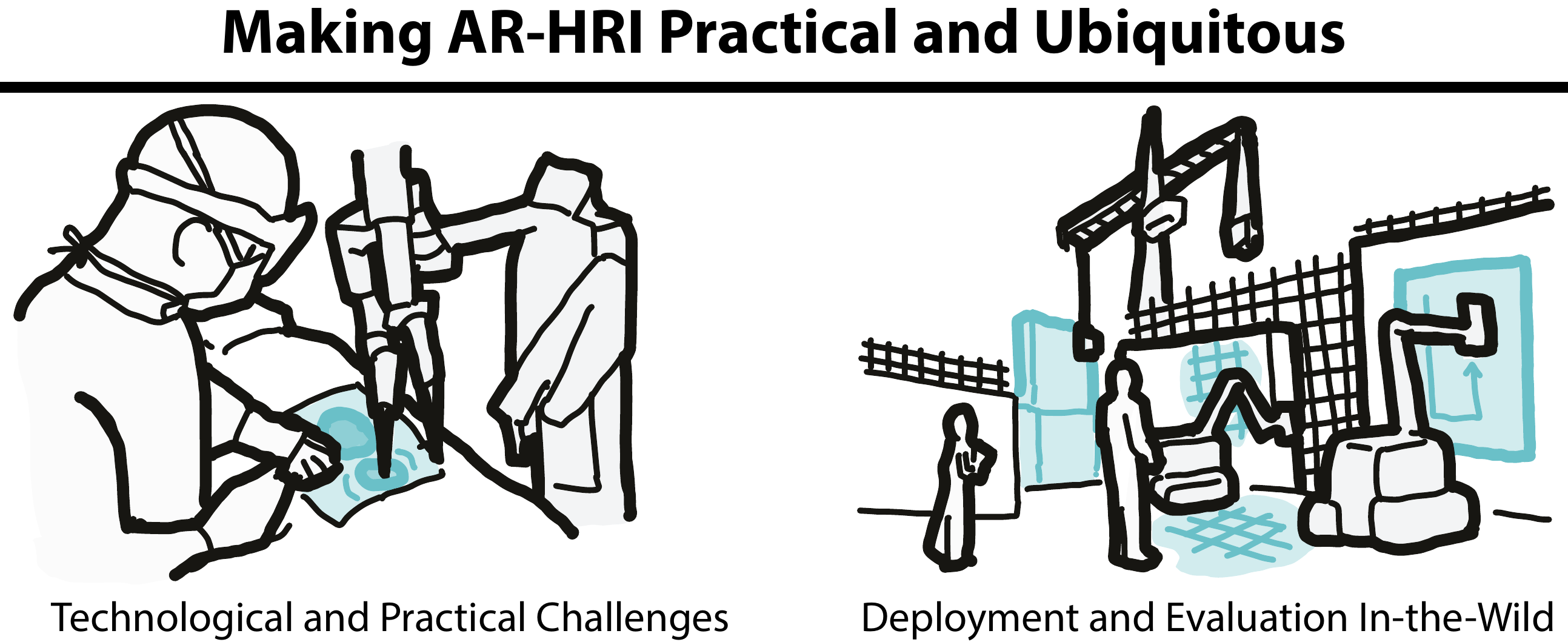}
\caption{Opportunity-1: Making AR-HRI Practical and Ubiquitous. Left: Improvement of display and tracking technologies would broaden practical applications like robotic surgery. Right: Deployment in-the-wild outside the research lab, such as construction sites, would benefit from user-centered design.}
\label{fig:future-opportunity-1}
\end{figure}
}

\newcommand{\FigureOpportunityTwo}{
\begin{figure}[h!]
\centering
\includegraphics[width=\FigureHalfWidth]{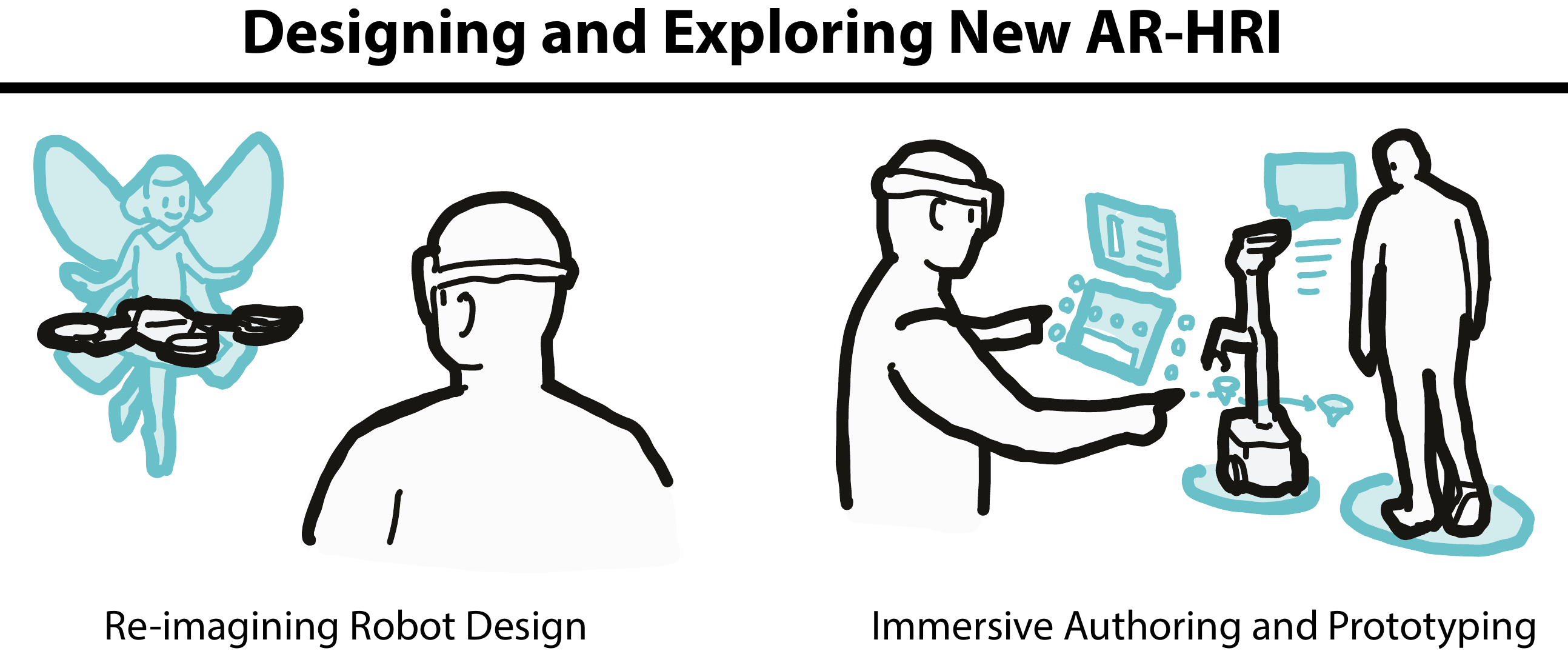}
\caption{Opportunity-2: Designing and Exploring New AR-HRI. Left: AR-HRI can open up a new opportunity for unconventional robot design like fairy or fictional characters with the power of AR. Right: Immersive authoring tools allow us to prototype interactions through direct manipulation within AR.}
\label{fig:future-opportunity-2}
\end{figure}
}

\newcommand{\FigureOpportunityThree}{
\begin{figure}[h!]
\centering
\includegraphics[width=\FigureHalfWidth]{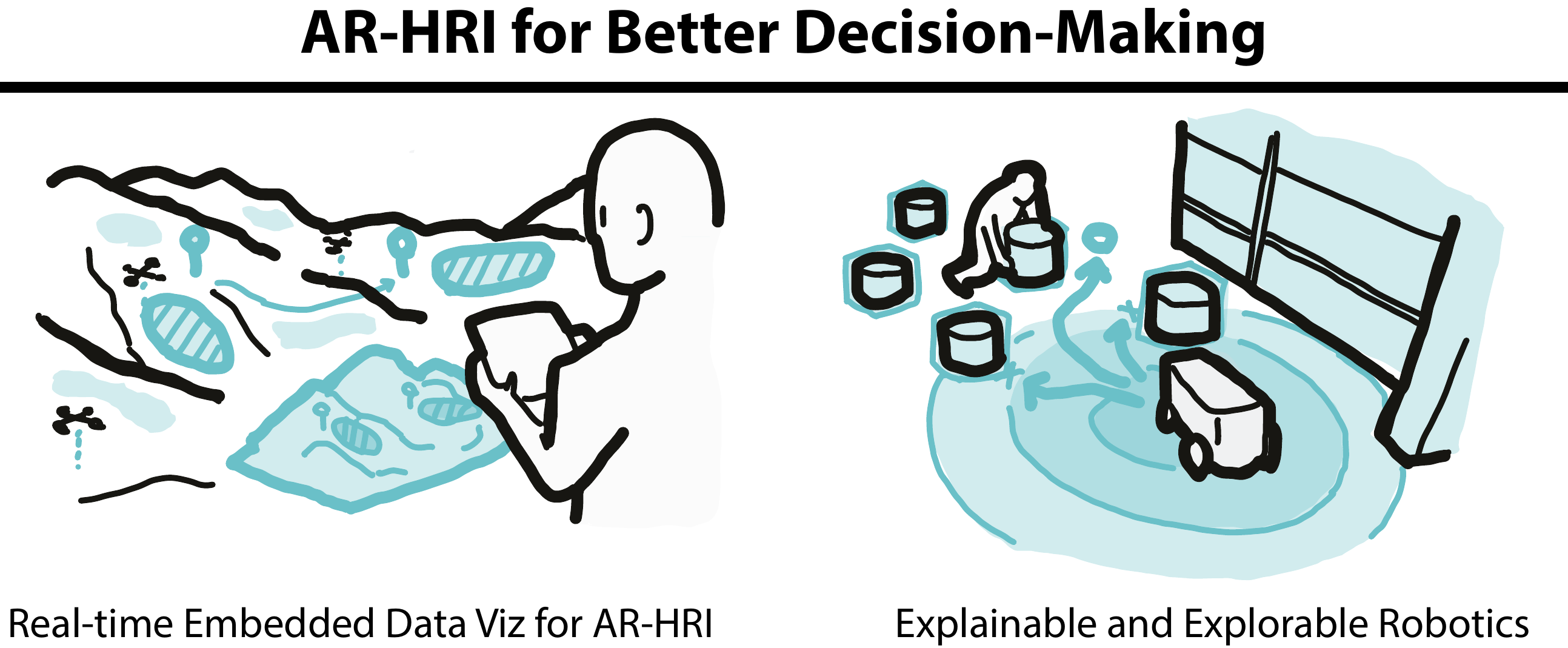}
\caption{Opportunity-3: AR-HRI for Better Decision-Making. Left: Real-time embedded and immersive visualizations help an operator's decision-making in drone navigation for search-and-rescue. Right: AR-based explainable robotics enables the user to understand a robot's path planning behavior through physical explorations.}
\label{fig:future-opportunity-3}
\end{figure}
}

\newcommand{\FigureOpportunityFour}{
\begin{figure}[h!]
\centering
\includegraphics[width=\FigureHalfWidth]{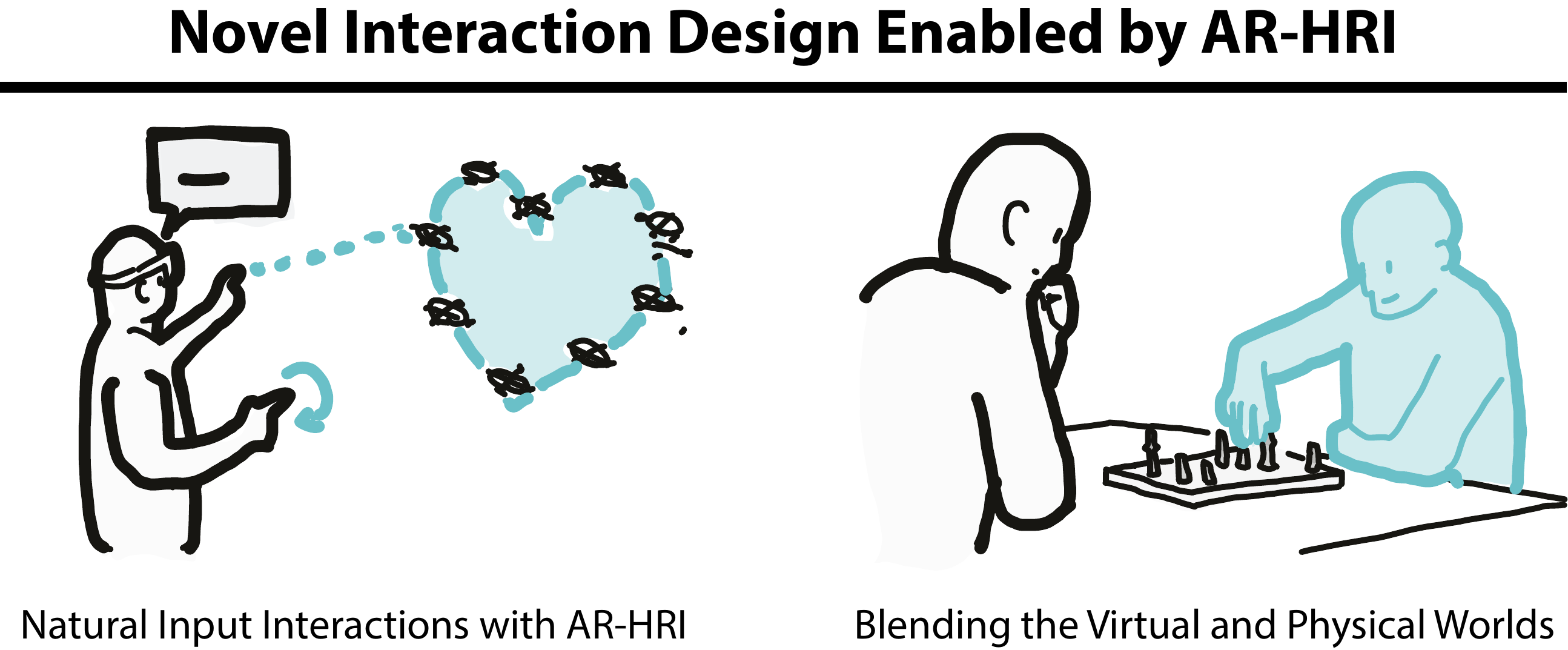}
\caption{Opportunity-4: Novel Interaction Design enabled by AR-HRI. Left: The user can interact with a swarm of drones with an expressive two-handed gesture like a mid-air drawing. Right: Programmable physical actuation and reconfiguration enable us to further blend the virtual and physical worlds, like a virtual remote user can ``physically'' move a chess piece with tabletop actuation.}
\label{fig:future-opportunity-4}
\end{figure}
}

%% file: 0-abstract.tex
\begin{abstract}
This paper contributes to a taxonomy of {\it augmented reality and robotics} based on a survey of 460 research papers. Augmented and mixed reality (AR/MR) have emerged as a new way to enhance human-robot interaction (HRI) and robotic interfaces (e.g., actuated and shape-changing interfaces). Recently, an increasing number of studies in HCI, HRI, and robotics have demonstrated how AR enables better interactions between people and robots. However, often research remains focused on individual explorations and key design strategies, and research questions are rarely analyzed systematically. In this paper, we synthesize and categorize this research field in the following dimensions: 1) approaches to augmenting reality; 2) characteristics of robots; 3) purposes and benefits; 4) classification of presented information; 5) design components and strategies for visual augmentation; 6) interaction techniques and modalities; 7) application domains; and 8) evaluation strategies. We formulate key challenges and opportunities to guide and inform future research in AR and robotics.
\end{abstract}

\ccsdesc[500]{Human-centered computing~Mixed / augmented reality}
\ccsdesc[500]{Computer systems organization~Robotics}

\keywords{survey; augmented reality; mixed reality; robotics; human-robot interaction; actuated tangible interfaces; shape-changing interfaces}

%% file: 1-introduction.tex
\ifdouble
\FigureScreenShots
\fi

\section{Introduction}
As robots become more ubiquitous, designing the best possible interaction between people and robots is becoming increasingly important.
Traditionally, interaction with robots often relies on the robot's internal physical or visual feedback capabilities, such as robots' movements~\cite{dragan2013legibility, zhou2017expressive, mainprice2010planning, szafir2014communication}, gestural motion~\cite{huang2013modeling, chidambaram2012designing, okuno2009providing}, gaze outputs~\cite{andrist2012designing,mutlu6modeling,kalegina2018characterizing,argyle1976gaze}, physical transformation~\cite{hedayati2020pufferbot}, or visual feedback through lights~\cite{szafir2015communicating,song2018bioluminescence,baraka2016expressive, cha2016using} or small displays~\cite{fortmann2013make, hegarty2010thinking, vogel2004interactive}. 
However, such modalities have several key limitations. For example, the robot's form factor cannot be easily modified on demand, thus it is often difficult to provide expressive physical feedback that goes beyond internal capabilities~\cite{walker2018communicating}.
While visual feedback such as lights or displays can be more flexible, the expression of such visual outputs is still bound to the fixed physical design of the robot.
For example, it can be challenging to present expressive information given the fixed size of a small display, where it cannot show the data or information associated with the physical space that is situated outside the screen.
Augmented reality (AR) interfaces promise to address these challenges, as AR enables us to design expressive visual feedback without many of the constraints of physical reality.
In addition, AR can present visual feedback in one's line of sight, tightly coupled with the physical interaction space, which reduces the user's cognitive load when switching the context and attention between the robot and an external display.
Recent advances in AR opened up exciting new opportunities for human-robot interaction research, and over the last decades, an increasing number of works have started investigating how AR can be integrated into robotics to augment their inherent visual and physical output capabilities.
However, often these research projects are individual explorations, and key design strategies, common practices, and open research questions in AR and robotics research are rarely analyzed systematically, especially from an interaction design perspective.
With the recent proliferation of this research field, we see a need to synthesize the existing works to facilitate further advances in both HCI and robotics communities. 

In this paper, we review a corpus of 460 papers to synthesize the taxonomy for AR and robotics research. In particular, we synthesized the research field into the following design space dimensions (with a brief visual summary in Figure~\ref{fig:teaser}):
1) approaches to augmenting reality for HRI; 2) characteristics of augmented robots; 3) purposes and benefits of the use of AR; 4) classification of presented information; 5) design components and strategies for visual augmentation; 6) interaction techniques and modalities; 7) application domains; and 8) evaluation strategies.
Our goal is to provide a common ground and understanding for researchers in the field, which both includes AR-enhanced {\it human-robot interaction}~\cite{goodrich2008human} and {\it robotic user interfaces}~\cite{beckerle2019robotic, kim2017ubiswarm} research (such as actuated tangible~\cite{poupyrev2007actuation} and shape-changing interfaces~\cite{coelho2011shape, rasmussen2012shape, alexander2018grand}).
We envision this paper can help researchers situate their work within the large design space and explore novel interfaces for AR-enhanced human-robot interaction (AR-HRI). 
Furthermore, our taxonomy and detailed design space dimensions (together with the comprehensive index linking to related work) can help readers to more rapidly find practical AR-HRI techniques, which they can then use, iterate and evolve into their own future designs.
Finally, we formulate open research questions, challenges, and opportunities to guide and stimulate the research communities of HCI, HRI, and robotics. 

%% file: 2-scope.tex
\ifsingle
\FigureScreenShots
\fi

\section{Scope, Contributions, and Methodology}

\subsection{Scope and Definitions}

The topic covered by this paper is {\it ``robotic systems that utilize AR for interaction''}. 
In this section, we describe this scope in more detail and clarify what is included and what is not.

\subsubsection{Human-Robot Interaction and Robotic Interfaces}
{\it ``Robotic systems''} could take different forms---from traditional industrial robots to self-driving cars or actuated user interfaces.
In this paper, we do not limit the scope of robots and include any type of robotic or actuated systems that are designed to interact with people.
More specifically, our paper also covers {\it robotic interface}~\cite{beckerle2019robotic, kim2017ubiswarm} research.
Here, {\it robotic interfaces} refer to interfaces that use robots and/or actuated systems as a medium for human-computer interaction~\footnote{We only cover {\it internally} actuated systems but do not cover {\it externally} actuated systems, which actuate passive objects with external force~\cite{patten2007mechanical, pangaro2002actuated, suzuki2018reactile, morales2019leviprops}.}.
This includes actuated tangible interfaces~\cite{poupyrev2007actuation}, adaptive environments~\cite{sirkin2015mechanical, green2016architectural}, swarm user interfaces~\cite{le2016zooids}, and shape-changing interfaces~\cite{coelho2011shape, rasmussen2012shape, alexander2018grand}.


\subsubsection{AR vs VR}
Among HRI and robotic interface research, we specifically focus on AR, but not on VR.
In the robotics literature, VR has been used for many different purposes, such as interactive simulation~\cite{herdel2021drone, mahadevan2021grip, mahadevan2019av} or haptic environments~\cite{mcneely1993robotic, vonach2017vrrobot, suzuki2020roomshift}.
However, our focus is on visual augmentation \textit{in the real world} to enhance real robots in the physical space, thus we specifically investigate systems that uses AR for robotics.

\subsubsection{What is AR}
The definition of AR can also vary based on the context~\cite{speicher2019mixed}. For example, Azuma defines AR as \textit{``systems that have the following three characteristics:
1) combines real and virtual, 2) interactive in real time, 3) registered in 3D''}~\cite{azuma1997survey}. Milgram and Kishino also describe this with the reality-virtuality continuum~\cite{milgram1994taxonomy}.
More broadly, Bimber and Rasker~\cite{bimber2006modern} also discuss spatial augmented reality (SAR) as one of the categories in AR.
In this paper, we take AR as a broader scope and include any systems that augment physical objects or surroundings environments in the real world, regardless of the technology used.

\subsection{Contributions}
Augmented reality in the field of robotics has been the scope of other related review papers (e.g.,~\cite{makhataeva2020augmented, dianatfar2021review, qian2019review, green2008human}) that our taxonomy expands upon.
Most of these earlier papers reviewed key application use cases in the research field. For example, Makhataeva and Varol surveyed example applications of AR for robotics in a 5-year timeframe~\cite{makhataeva2020augmented} and Qian et al. reviewed AR applications for \textit{robotic surgery} in particular~\cite{qian2019review}.
From the HRI perspective, Green et al. provide a literature review research for collaborative HRI~\cite{green2008human}, which focuses in particular on collaboration through the means AR technologies. And more recently, human-robot interaction and VR/MR/AR (VAM-HRI) as also been the topic of workshops~\cite{williams2018virtual}.

Our taxonomy builds on and extends beyond these earlier reviews. In particular, we provide the following contributions.
First, we present a taxonomy with a novel set of \textbf{\textit{design space dimensions}}, providing a holistic view based on the different dimensions unifying the design space, with a focus on {\it interaction and visual augmentation design perspectives}. 
Second, our paper also systematically covers a \textbf{\textit{broader scope of HCI and HRI literature}}, including robotic, actuated, and shape-changing user interfaces.
This field is increasingly popular in the field of HCI, ~\cite{poupyrev2007actuation, coelho2011shape, rasmussen2012shape, alexander2018grand} but not well explored in terms of the combination with AR. 
By incorporating this research, our paper provides a more comprehensive view to position and design novel AR/MR interactions for robotic systems. 
Third, we also discuss \textbf{\textit{open research questions and opportunities}} that facilitate further research in this field. 
We believe that our taxonomy\,---\,with the design classifications and their insights, and the articulation of open research questions\,---\,will be invaluable tools for providing a common ground and understanding when designing AR/MR interfaces for HRI. This will help researchers identify or explore novel interactions. 
Finally, we also compiled a large corpus of research literature using our taxonomy as an \textbf{\textit{interactive website}}~\footnote{\url{https://ilab.ucalgary.ca/ar-and-robotics/}}, which can provide a more content-rich, up-to-date, and extensible literature review.
Inspired by similar attempts in personal fabrication~\cite{baudisch2017personal, personal-fab}, data physicalization~\cite{jansen2015opportunities, data-phys}, and material-based shape-changing interactions~\cite{qamar2018hci, morph-ui}, our website, along with this paper, could provide similar benefits to the broader community of both researchers and practitioners.

\ifdouble
\FigureApproach
\fi

\subsection{Methodology}

\subsubsection{Dataset and Inclusion Criteria}
To collect a representative set of AR and robotics papers, we conducted a systematic search in the ACM Digital Library, IEEE Xplore, MDPI, Springer, and Elsevier.
Our search terms include the combination of {\it ``augmented reality''} AND {\it ``robot''} in the title and/or author keywords since 2000.
We also searched for synonyms of each keyword, such as {\it ``mixed reality'', ``AR'', ``MR''} for augmented reality and {\it ``robotic'', ``actuated'', ``shape-changing''} for robot.
This gave us a total of 925 papers after removing duplicates.
Then, four authors individually looked at each paper to exclude out-of-scope papers, which, for example, only focus on AR-based tracking but not on visual augmentation, or were concept or position papers, etc.
After this process, we obtained 396 papers in total.
To complement this keyword search, we also identified an additional relevant 64 papers by leveraging the authors' expertise in HCI, HRI, and robotic interfaces.
By merging these papers, we finally selected a corpus of 460 papers for our literature review.

While our systematic compilation of this corpus provides an in-depth view into the research space, this set can not be a complete or exhaustive list in this domain. The boundaries and scope of our corpus may not be clear cut, and as with any selection of papers, there were many papers right at the boundaries of our inclusion/exclusion criteria. Nevertheless, our focus was on the development of a taxonomy and this corpus serves as a representative subset of the most relevant papers. We aim to address this inherent limitation of any taxonomy by making our coding and dataset open-source, available for others to iterate and expand upon.


\subsubsection{Analysis and Synthesis}
The dataset was analyzed through a multi-step process. One of the authors conducted open-coding on a small subset of our sample to identify a first approximation of the dimensions and categories within the design space.
Next, all authors reflected upon the initial design space classification to discuss the consistency and comprehensiveness of the categorization methods, where then categories were merged, expanded, and removed.
Next, three other co-authors performed systematic coding with individual tagging for  categorization of the complete dataset.
Finally, we reflected upon the individual tagging to resolve the discrepancies to obtain the final coding results.

In the following sections, we present our results and findings of this classification by using \textit{color-coded text and figures}. 
We provide a list of \textit{key citations} directly within the figures, with the goal of facilitating lookup of relevant papers within each dimension and all of the corresponding sub-categories. Furthermore, in the appendix of this paper we included several tables with a complete compilation of all citations and count of the papers in our corpus that fall within each of the categories and sub-categories of the design space\,---\,which we hope will help researchers to more easily find relevant papers (e.g., finding all papers that use AR for \textit{"improving safety"} with robots, \textit{"augment surroundings"} of robots, or provide visual feedback of \textit{"paths and trajectories"}).   



%% file: 3-approach.tex
\coloredSection{color1}{Approaches to Augmenting Reality in Robotics}
In this section, we discuss the different approaches to augmenting reality in robotics (\textcolor{color1}{ Figure~\ref{fig:approaches}}).
To classify how to augment reality, we propose to categorize based on two dimensions: 
First, we categorize the approaches based on the \textbf{placement of the augmented reality hardware} (i.e., where the optical path is overridden with digital information). 
For our purpose, we adopt and extend Bimber and Raskar's~\cite{bimber2006modern} classification in the context of robotics research.
Here, we propose three different locations: 1) {\it on-body}, 2) {\it on-environment}, and 3) {\it on-robot}.
Second, we classify based on the {\bf target location of visual augmentation}, i.e., where is augmented.
We can categorize this based on 1) \textcolor{color1}{{\it augmenting robots}} or 2) \textcolor{color2}{{\it augmenting surroundings}}. 
Given these two dimensions, we can map the existing works into the design space (\textcolor{color1}{Figure~\ref{fig:approaches} Right}). Walker et al.~\cite{walker2018communicating} include augmenting user interface (UI) as another category. Since the research that has been done in this area can be roughly considered augmenting the environment, we decided to not include it as a separate category.

\ifsingle
\FigureApproach
\fi

\approach{1}{Augment Robots}{color1}
AR is used to augment robots themselves by overlaying or anchoring additional information on top of the robots (\textcolor{color1}{Figure~\ref{fig:approaches} Top}).

\approachsub{On-Body}{color1}
The first category augments robots through on-body AR devices.
This can be either 1) head-mounted displays (HMD)~\cite{rosen2020communicating, jones2020vroom, walker2018communicating} or 2) mobile AR interfaces~\cite{chen2021pinpointfly, kasahara2013extouch}.
For example, VRoom~\cite{jones2020vroom, jones2021belonging} augments the telepresence robot's appearance by overlaying a remote user.
Similarly, Young et al.~\cite{young2007robot} demonstrated adding an animated face onto a Roomba robot to show an expressive emotion on mobile AR devices.

\approachsub{On-Environment}{color1}
The second category augments robots with devices embedded in the surrounding environment.
Technologies often used with this approach include 1) environment-attached projectors~\cite{andersen2016projecting} or 2) see-through displays~\cite{leithinger2013sublimate}.
For example, DroneSAR~\cite{darbar2019dronesar} also shows how we can augment the drone itself with projection mapping.
Showing the overlaid information on top of robotic interfaces can also fall into this category.
Similarly, shape-shifting walls~\cite{takashima2016study} or handheld shape-changing interfaces~\cite{lindlbauer2016combining, roudaut2013morphees} are also directly augmented with the overlaid animation of information. 

\approachsub{On-Robot}{color1}
In the third category, the robots augment their own appearance, which is unique in AR and robotics research, compared to Bimber and Raskar's taxonomy~\cite{bimber2006modern}.
For example, Furhat~\cite{al2012furhat} animates a face with a back-projected robot head, so that the robot can augment its own face without an external AR device.
The common technologies used are robot-attached projectors~\cite{tobita2011floating, suzuki2016gushed}, which augments itself by using its own body as a screen.
Alternatively, robot-attached displays can also fall into this category~\cite{villanueva2021robotar, yamada2017isphere}.

\approach{2}{Augment Surroundings}{color2}
Alternatively, AR is also used to augment the surroundings of the robots.
Here, the surroundings include 1) surrounding mid-air 3D {\it space}, 2) surrounding physical {\it objects}, or 3) surrounding physical {\it environments}, such as wall, floor, ceiling, etc (\textcolor{color2}{Figure~\ref{fig:approaches} Bottom}). 

\approachsub{On-Body}{color2}
Similarly, this category augments robots' surroundings through 1) HMD~\cite{rosen2020communicating, walker2018communicating}, 2) mobile AR devices~\cite{chen2021pinpointfly}, or 3) handheld projector~\cite{hiraki2019navigatorch}.
One benefit of HMD or mobile AR devices is an expressive rendering capability enabled by leveraging 3D graphics and spatial scene understanding.
For example, Drone Augmented Human Vision~\cite{erat2018drone} uses HMD-based AR to change the appearance of the wall for remote control of drones.
RoMA~\cite{peng2018roma} uses HMD for overlaying the interactive 3D models on a robotic 3D printer. 

\approachsub{On-Environment}{color2}
In contrast to HMD or handheld devices, the on-environment approach allows much easier ways to share the AR experiences with co-located users.
Augmentation can be done through 1) projection mapping~\cite{suzuki2019shapebots} or 2) surface displays~\cite{guo2009touch}.
For example, Touch and Toys~\cite{guo2009touch} leverage a large surface display to show additional information in the surroundings of robots.
Andersen et al.~\cite{andersen2016projecting} investigates the use of projection mapping to highlight or augment surrounding objects to communicate the robot's intentions.
While it allows the shared content for multiple people, the drawback of this approach is a fixed location due to the requirements of installed-equipment, which may limit the flexibility and mobility for outdoor scenarios.

\approachsub{On-Robot}{color2}
In this category, the robots themselves augment the surrounding environments. 
We identified that the common approach is to utilize the robot-attached projector to augment surrounding physical environments~\cite{kasetani2015projection, scheible2013displaydrone}.
For example, Kasetani et al.~\cite{kasetani2015projection} attach a projector to a mobile robot to make a  self-propelled projector for ubiquitous displays. 
Moreover, DisplayDrone~\cite{scheible2013displaydrone} shows a projected image onto the surrounding walls for on-demand displays. 
The main benefit of this approach is that the user does not require any on-body or environment-instrumented devices, thus it enables mobile, flexible, and deployable experiences for different situations.

%% file: 4-robots.tex
\ifdouble
\FigureRobot
\FigurePurpose
\fi

\coloredSection{color3}{Characteristics of Augmented Robots}
Next, we classify research projects based on the characteristics of augmented robots. Possible design space dimensions span 1) the form factor of robots, 2) the relationship between the users and robots, 3) size and scale of the robots, and 4) proximity for interactions (\textcolor{color3}{Figure~\ref{fig:robots}}).

\ifsingle
\FigureRobot
\fi

\robot{1}{Form Factor}
This category includes the types of robots that have been investigated in the literature.
The form factor of robots include: robotic arms~\cite{qian2019augmented, peng2018roma}, drones~\cite{hedayati2018improving, cauchard2019drone}, mobile robots~\cite{wu2020mixed, jones2020vroom}, humanoid robots~\cite{tran2020exploring, li2019augmented}, vehicles~\cite{ochiai2011homunculus, morita2020extension}, actuated objects~\cite{gronbaek2020kirigamitable, urbani2018exploring}, the combination of multiple form factors~\cite{hashimoto2011touchme}, and other types such as fabrication machines~\cite{mueller2012interactive, yamaoka2016mirageprinter}.

\robot{2}{Relationship}
Research also explores different people-to-robot relationships. 
In the most common case, one person interacts with a single robot (1:1), but the existing research also explores a situation where one person interacts with multiple robots (1:m). 
AR for swarm robots falls into this category~\cite{suzuki2019shapebots, hiraki2018phygital, le2016zooids, ozgur2017cellulo}.
On the other hand, collaborative robots require multiple people to interact with a single robotic interface (n:1)~\cite{takashima2013transformtable} or a swarm of robots (n:m)~\cite{ozgur2017cellulo}. 

\robot{3}{Scale}
Augmented robots are of different sizes, along a spectrum from \textit{small} to \textit{large}: from a small handheld-scale which can be grasped with a single hand~\cite{suzuki2019shapebots}, tabletop-scale which can fit onto the table~\cite{linder2010luminar}, and body-scale which is about the same size as human bodies like industrial robotic arms~\cite{arevalo2021assisting, maly2016augmented}.
Large-scale robots are possible, such as vehicles~\cite{ochiai2011homunculus, mercedes-f15, jaguar-rover} or even building construction robots. 

\robot{4}{Proximity}
Proximity refers to the distance between the user and robots when interaction happens. 
Interactions can vary across the dimension of proximity, from \textit{near} to \textit{far}. 
The proximity can be classified as the spectrum between 1) {\it co-located} or 2) {\it remote}.
The proximity of the robots can influence whether the robots are directly touchable~\cite{nowacka2013touchbugs, krzywinski2009robotable} or situated in distance~\cite{darbar2019dronesar}.
It can also affect how to augment reality, based on whether the robots are visible to the user~\cite{hedayati2018improving} or out-of-sight for remote interaction~\cite{erat2018drone}.

%% file: 5-purpose.tex
\ifdouble
\FigureInformation 
\fi

\coloredSection{color4}{Purposes and Benefits of Visual Augmentation}
Visual augmentation has many benefits for effective human-robot interaction. 
In this section, we categorize the purposes of why visual augmentation is used in robotics research. 
On a higher level, purposes and benefits can be largely categorized as 1) for programming and control, and 2) for understanding, interpretation, and communications (\textcolor{color4}{Figure~\ref{fig:purposes}}).

\ifsingle
\FigurePurpose
\fi

\purpose{1}{Facilitate Programming}
First, the AR interface provides a powerful assistant to facilitate programming robots~\cite{bambusek2019combining}.
One way to facilitate the programming is to {\it simulate programmed behaviors}~\cite{guhl2017concept}, which has been explored since early 1990s~\cite{milgram1993applications, azuma1997survey, kim1996virtual}.
For example, GhostAR~\cite{cao2019ghostar} shows the trajectory of robots to help the user see how the robots will behave. 
Such visual simulation helps the user to program the robots in industry applications~\cite{quintero2018robot} or home automation~\cite{liu2011roboshop}.
Another aspect of programming assistance is to {\it directly map with the real world}.
Robot programming often involves interaction with real-world objects, and going back and forth between physical and virtual worlds is tedious and time-consuming.
AR interfaces allow the user to directly indicate objects or locations in the physical world.
For example, Gong et al.~\cite{gong2019projection} utilizes projection-based AR to support the programming of grasping tasks.

\purpose{2}{Support Real-time Control and Navigation}
Similar to the previous category, AR interfaces facilitate the control, navigation, and teleoperation of the robot.
In contrast to programming the behaviors, this category focuses on the {\it real-time} operation of the robot, either remote or co-located. 
For example, exTouch~\cite{kasahara2013extouch} and PinpointFly~\cite{chen2021pinpointfly} allows the user to interactively control robots with the visual feedback on a touch screen.
AR interfaces also support showing additional information or parameters related to the navigation and control.
For example, a world-in-miniature of the physical world~\cite{boeing-uav} or real-time camera view~\cite{hedayati2018improving} is used to support remote navigation of drones.

\purpose{3}{Improve Safety}
By leveraging visual augmentation, AR/MR interfaces can improve safety awareness when interacting with robots. 
For example, Safety Aura Visualization~\cite{makhataeva2019safety} explores spatial color mapping to indicate the safe and dangerous zones. 
Virtual barriers in AR~\cite{hoang2021virtual, chan2018virtual} help the user avoid unexpected collisions with the robots.

\purpose{4}{Communicate Intent}
AR interfaces can also help to communicate the robot's intention to the user through spatial information.
For example, Walker et al. show that the AR representations can better communicate the drone's intent through the experiments using three different designs~\cite{walker2018communicating}. 
Similarly, Rosen et al. reveal that the AR visualization can better present the robotic arm's intent through the spatial trajectory, compared to the traditional interfaces~\cite{rosen2020communicating}. 
AR interfaces can be also used to indicate the state of robot manipulation such as indicating warning or completion of the task~\cite{andersen2016projecting} or communicating intent with passersby or pedestrians for wheelchairs~\cite{watanabe2015communicating} or self-driving cars~\cite{ochiai2011homunculus}. 

\purpose{5}{Increase the Expressiveness}
Finally, AR can also be used to augment the robot's expression~\cite{hololens-robot}.
For example, Groechel et al.~\cite{groechel2019using} uses an AR view to provide virtual arms to a social robot (e.g., Kuri Robot) to enhance the social expressions when communicating with the users.
Examples include adding facial expressions~\cite{young2007robot}, overlaying remote users~\cite{jones2021belonging, siu2018investigating}, and interactive content~\cite{darbar2019dronesar} onto robots.
AR is a helpful medium to increase the expressiveness of shape-changing interfaces~\cite{lindlbauer2016combining}.
For example, Sublimate~\cite{leithinger2013sublimate} or inFORM~\cite{follmer2013inform} uses see-through display or projection mapping to provide a virtual surface on a shape display.

%% file: 6-information.tex
\coloredSection{color5}{Classification of Presented Information}
This section summarizes types of information presented in AR interfaces. 
The categories we identified include 1) robot's internal information, 2) external information about the environment, 3) plan and activity, and 4) supplemental content (\textcolor{color5}{Figure~\ref{fig:information}}).

\ifsingle
\FigureInformation
\fi

\ifdouble
\FigureDesign
\fi

\information{1}{Robot's Internal Information}
The first category is the robot's internal information. 
This can include 1) robot's internal status, 2) robot's software and hardware condition, 3) robot's internal functionality and capability.
Examples include the robot's emotional state for social interaction~\cite{groechel2019using, young2007robot}, a warning sign when the user's program is wrong~\cite{liu2018interactive}, the drone's current information such as altitude, flight mode, flight status, and dilution of precision~\cite{papachristos2016augmented, aleotti2017detection}, and the robot's reachable region to indicate safe and dangerous zones~\cite{makhataeva2019safety}.
Showing the robot's hardware components is also included in this category. 
For example, showing or highlighting physical parts of the robot for maintenance~\cite{mourtzis2017augmented, maly2016augmented} is also classified as this category.

\information{2}{External Information about the Environment}
Another category is external information about the environment. 
This includes 1) sensor data from the internal or external sensors, 2) camera or video feed, 3) information about external objects, 4) depth map or 3D reconstructed scene of the environment. 
Examples include camera feeds for remote drone operations~\cite{hedayati2018improving}, the world in miniature of the environment~\cite{boeing-uav}, sensor stream data of the environment~\cite{aleotti2017detection}, visualization of obstacles~\cite{liu2011roboshop}, a local cost map for search task~\cite{muhammad2019creating}, a 3D reconstructed view of the environment~\cite{papachristos2016augmented, erat2018drone}, a warning sign projected onto an object that indicates the robot's intention~\cite{andersen2016projecting}, visual feedback about the localization of the robot~\cite{wu2020mixed}, and position and label of objects for grasping tasks~\cite{gradmann2018augmented}.
Such embedded external information improves the situation awareness and comprehension of the task, especially for real-time control and navigation.

\information{3}{Plan and Activity}
The previous two categories focus on the {\it current} information, but plan and activity are related to {\it future} information about the robot's behavior. 
This includes 1) a plan of the robot's motion and behavior, 2) simulation results of the programmed behavior, 3) visualization of a target and goal, 4) progress of the current task.
Examples include the future trajectory of the drone~\cite{walker2018communicating}, the direction of the mobile robots or vehicles~\cite{huy2017see, ochiai2011homunculus}, a highlight of the object the robot is about to grasp~\cite{bambusek2019combining}, the location of the robot's target position~\cite{yuan2019human}, and a simulation of the programmed robotic arm's motion and behavior~\cite{rosen2020communicating}.
This type of information helps the user better understand and expect the robot's behavior and intention.

\information{4}{Supplemental Content}
Finally, AR is also used to show supplemental content for expressive interaction, such as showing interactive content on robots or background images for their surroundings. 
Examples include a holographic remote user for remote collaboration and telepresence~\cite{jones2020vroom, siu2018investigating}, a visual scene for games and entertainment~\cite{piumatti2017spatial, robert2012blended}, an overlaid animation or visual content for shape-changing interfaces~\cite{lindlbauer2016combining, leithinger2013sublimate}, showing the menu for available actions~\cite{cauchard2019drone, arevalo2021assisting}, and aided color coding or background for dynamic data physicalization~\cite{suzuki2019shapebots, follmer2013inform}.

%% file: 7-design.tex
\coloredSection{color6}{Design Components and Strategies for Visual Augmentation}
Different from the previous section that discusses {\it what} to show in AR, this section focuses on {\it how} to show AR content.
To this end, we classify common design practices across the existing visual augmentation examples.
At a higher level, we identified the following design strategies and components: 1) UIs and widgets, 2) spatial references and visualizations, and 3) embedded visual effects (\textcolor{color6}{Figure~\ref{fig:design}}).

\ifsingle
\FigureDesign
\fi

\design{1}{UIs and Widgets}
UIs and widgets are a common design practice in AR for robotics to help the user see, understand, and interact with the information related to robots (\textcolor{color6}{Figure~\ref{fig:design} Top}).

\designsub{Menus} 
The menu is often used in mixed reality interfaces for human-robot interaction~\cite{ostanin2020human, gao2019pati, sugimoto2011robotable2}. 
The menu helps the user to see and select the available options~\cite{ostanin2018interactive}. 
The user can also control or communicate with robots through a menu and gestural interaction~\cite{cauchard2019drone}.

\designsub{Information Panels}
Information panels show the robot's internal or external status as floating windows~\cite{urbani2018exploring} with either textual or visual representations.
Textual information can be effective to present precise information such as the current altitude of the drone~\cite{aleotti2017detection} or the measured length~\cite{darbar2019dronesar}. 
More complex visual information can also shown such as a network graph of the current task and program~\cite{liu2018interactive}. 

\designsub{Labels and Annotations}
Labels and annotations are used to show information about the object.
Also, they are used to annotate objects~\cite{darbar2019dronesar}.

\designsub{Controls and Handles}
Controls and handles are another user interface example.
They allow the user to control robots through a virtual handle~\cite{hashimoto2011touchme}.
Also, AR can show the control value surrounding the robot~\cite{pedersen2011tangible}.

\designsub{Monitors and Displays}
Monitor or displays help the user to situate themselves in the remote environment~\cite{villanueva2021robotar}.
Camera monitors allow the user to better navigate the drone for inspection or aerial photography tasks~\cite{hedayati2018improving}.
The camera feed can be also combined with the real-time 3D reconstruction~\cite{papachristos2016augmented}.
In contrast, monitor or display are also used to display spatially registered content in the surrounding environment ~\cite{scheible2013displaydrone} or on top of the robot~\cite{urbani2018exploring}

\design{2}{Spatial References and Visualizations}
Spatial references and visualizations are a technique used to overlay data spatially. 
Similar to embedded visualizations~\cite{willett2016embedded}, this design can directly embed data on top of their corresponding physical referents.
The representation can be from a simple graphical element, such as points (0D), paths (1D), or areas (2D/3D), to more complex visualizations like color maps (\textcolor{color6}{Figure~\ref{fig:design} Middle}).

\designsub{Points and Locations}
Points are used to visualize a specific location in AR.
These points can be used to highlight a landmark~\cite{fuste2020kinetic}, target location~\cite{yuan2019human}, or way point~\cite{walker2019robot}, which is associated to the geo-spatial information.
Additionally, points can be used as a control or anchor point to manipulate virtual objects or boundaries~\cite{ostanin2018interactive}. 

\designsub{Paths and Trajectories}
Similarly, paths and trajectories are another common approaches to represent spatial references as lines~\cite{stadler2016augmented, quintero2018robot, walker2018communicating}.
For example, paths are commonly used to visualize the expected behaviors for real-time or programmed control~\cite{zollmann2014flyar, chen2021pinpointfly, chen2019pinpointfly}. 
By combining the interactive points, the user can modify these paths by adding, editing, or deleting the way points~\cite{walker2019robot}.

\designsub{Areas and Boundaries}
Areas and boundaries are used to highlight specific regions of the physical environment.
They can visualize a virtual bounding box for safety purposes~\cite{chan2018virtual, hoang2021virtual} or highlight a region to show the robot's intent~\cite{andersen2016projecting, dinh2017augmented}.
Alternatively, the areas and boundaries are also visualized as a group of objects or robots~\cite{ishii2009designing}. 
Some research projects also demonstrated the use of interactive sketching for specifying the boundaries in home automation~\cite{liu2011roboshop, ishii2009designing}. 

\designsub{Other Visualizations}
Spatial visualizations can also take more complex and expressive forms. 
For example, spatial color/heat map visualization can indicate the safe and danger zones in the workspace, based on the robot's reachable areas~\cite{makhataeva2019safety}.
Alternatively, a force map visualizes the field of force to provide visual affordance for the robot's control~\cite{kato2009multi, hiraki2019navigatorch, hiraki2018phygital}. 

\design{3}{Embedded Visual Effects}
Embedded visual effects refer to graphical content directly embedded in the real world.
In contrast to spatial visualization, embedded visualization does not need to encode data. 
Common embedded visual effects are 1) anthropomorphic effects, 2) virtual replica, and 3) texture mapping of physical objects (\textcolor{color6}{Figure~\ref{fig:design} Bottom}).

\designsub{Anthropomorphic Effects}
Anthropomorphic effects are visual augmentations that render human-inspired graphics.
Such design can add an interactive effect of 1) a robot's body~\cite{hololens-robot}, such as arms~\cite{groechel2019using} and eyes~\cite{walker2018communicating}, 2) faces and facial expressions~\cite{al2012furhat, young2007robot, hiroi2010evaluation}, 3) a human-avatar~\cite{jones2021belonging, siu2018investigating, xiao2013mirrorfugue}, or 4) character animation~\cite{aoki2005kobito, xiao2016andantino}, on top of the robots.
For example, it can augment the robot's face by animated facial expression with realistic images~\cite{al2012furhat} or cartoon-like animation~\cite{young2007robot}, which can improve the social expression of the robots~\cite{groechel2019using} and engage more interaction~\cite{aoki2005kobito, xiao2016andantino}.
In addition to augmenting a robot's body, it can also show the image of a real person to facilitate remote communication~\cite{jones2020vroom, tobita2011floating, siu2018investigating, xiao2013mirrorfugue}.

\designsub{Virtual Replica and Ghost Effects}
A virtual replica is a 3D rendering of robots, objects, or external environments.
By combining with spatial references, a virtual replica is helpful to visualize the simulated behaviors~\cite{hashimoto2011touchme, rosen2020communicating, chen2021pinpointfly, walker2019robot, zollmann2014flyar}.
By rendering multiple virtual replicas, the system can also show the ghost effect with a series of semi-transparent replica~\cite{cao2019ghostar, rosen2020communicating}.
In addition, a replica of external objects or environments is also used to facilitate co-located programming~\cite{bambusek2019combining, arevalo2020there} or real-time navigation in the hidden space~\cite{erat2018drone}. 
Also, a miniaturized replica of the environment (i.e., the world in miniature) helps drone navigation~\cite{boeing-uav}.

\designsub{Texture Mapping Effects based on Shape} 
Finally, texture mapping overlays interactive content onto physical objects to increase expressiveness. 
This technique is often used to enhance shape-changing interfaces and displays~\cite{follmer2013inform, nakagaki2016materiable, roudaut2013morphees, hirai2018xslate}, such as overlaying terrain~\cite{leithinger2010relief, leithinger2011direct}, landscape~\cite{everitt2017polysurface}, animated game elements~\cite{lindlbauer2016combining, takashima2016study}, colored texture~\cite{nakagaki2016materiable}, or NURBS (Non-Uniform Rational Basis Spline) surface effects~\cite{leithinger2013sublimate}.
Texture effects can also augment the surrounding background of the robot.
For example, by overlaying the background texture onto the surrounding walls or surfaces, AR can contextualize the robots with the background of an immersive educational game~\cite{robert2012blended, robert2010exploring}, a visual map~\cite{suzuki2019shapebots, takashima2016study, pedersen2011tangible}, or a solar system~\cite{ozgur2017cellulo}.

%% file: 8-interactions.tex
\coloredSection{color7}{Interactions}

\interaction{1}{Level of Interactivity}{fig:interactivity}
In this section, we survey the interactions in AR and robotics research. The first dimension is level of interactivity (\textcolor{color7}{Figure~\ref{fig:interactivity}}).

\interactionsub{No Interaction (Only Output)}
In this category, the system uses AR solely for visual output and disregards user input~\cite{young2007robot, groechel2019using, suzuki2016gushed, tobita2011floating, scheible2013displaydrone, rosen2020communicating, walker2018communicating, lingamaneni2017dronecast, papachristos2016augmented}.
Examples include visualization of the robot's motion or capability~\cite{walker2018communicating, rosen2020communicating, makhataeva2019safety}, but these systems often focus on visual outputs, independent of the user's action. 

\FigureInteractivity

\ifdouble
\FigureInteraction
\fi

\interactionsub{Implicit Interaction}
Implicit interaction takes the user's implicit motion as input, such as the user's position or proximity to the robot~\cite{watanabe2015communicating}. Sometimes, the user may not necessarily realize the association between their actions and effects, but the robots respond implicitly to the users’ physical movements (e.g., approaching to the robot).

\interactionsub{Explicit and Indirect Manipulation}
Indirect manipulation is the user's input through remote manipulation without any physical contact.
The interaction can take place through pointing out objects~\cite{ostanin2018interactive}, selecting and drawing~\cite{ishii2009designing}, or explicitly determining actions with body motion (e.g., changing the setting in a virtual  menu~\cite{cauchard2019drone})


\interactionsub{Explicit and Direct Physical Manipulation}
Finally, this category involves the user's direct touch inputs with their hands or bodies. 
The user can physically interact with the robots through embodied body interaction~\cite{robert2012blended}.
Several interaction techniques utilize the deformation of objects or robots~\cite{leithinger2013sublimate}, grasping and manipulating~\cite{guo2009touch}, or physically demonstrating~\cite{luebbers2019augmented}.

\interaction{2}{Interaction Modalities}{fig:interactions}
Next, we synthesize categories based on the interaction modalities (\textcolor{color7}{Figure~\ref{fig:interactions}}). 

\interactionsub{Tangible}
The user can interact with robots by changing the shape or by physically deforming the object~\cite{leithinger2013sublimate, lindlbauer2016combining}, moving robots by grasping and moving tangible objects~\cite{guo2009touch, pedersen2011tangible}, or controlling robots by grasping and manipulating robots themselves~\cite{ozgur2017cellulo}. 

\interactionsub{Touch}
Touch interactions often involve the touch screen of mobiles, tablets, or other interactive surfaces.
The user can interact with robots by dragging or drawing on a tablet~\cite{kasahara2013extouch, chen2021pinpointfly, kato2009multi}, touching and pointing the target position~\cite{guo2009touch}, and manipulating virtual menus on a smartphone~\cite{cao2019v}.
The touch interaction is particularly useful when requiring precise input for controlling~\cite{hashimoto2011touchme, gradmann2018augmented} or programming the robot's motion~\cite{fuste2020kinetic, stadler2016augmented}.

\ifsingle
\FigureInteraction
\fi

\interactionsub{Pointer and Controller}
The pointer and controller allow the user to manipulate robots through spatial interaction or device action. 
Since the controller provides tactile feedback, it reduces the effort to manipulate robots~\cite{hedayati2018improving}.
While many controller inputs are explicit interactions~\cite{wistort2011tofudraw, ishii2009designing}, the user can also implicitly communicate with robots, such as designing a 3D virtual object with the pointer~\cite{peng2018roma}. 

\interactionsub{Spatial Gesture}
Spatial gestures are a common interaction modality for HMD-based interfaces~\cite{liu2018interactive, erat2018drone, arevalo2021assisting, ostanin2020human, cauchard2019drone, ochiai2011homunculus, bambusek2019combining, cha2018effects}.
With these kinds of gestures, users can manipulate virtual way points~\cite{quintero2018robot,ostanin2018interactive} or operate robots with a virtual menu~\cite{cauchard2019drone}.
The spatial gesture is also used to implicitly manipulate swarm robots through remote interaction~\cite{siu2018investigating}. 

\interactionsub{Gaze}
Gaze is often used to accompany the spatial gesture~\cite{yuan2019human, morita2020extension, quintero2018robot, ostanin2018interactive, liu2018interactive, erat2018drone, arevalo2021assisting}, such as when performing menu selection~\cite{arevalo2021assisting}.
But, some works investigate the gaze itself to control the robot by pointing out the location in 3D space~\cite{argyle1976gaze}.

\interactionsub{Voice}
Some research leveraged voice input to execute commands for the robot operation~\cite{arevalo2021assisting, qian2019augmented, hoang2021virtual, jones2020vroom}, especially in co-located settings.

\interactionsub{Proximity}
Finally, proximity is used as an implicit form of interaction to communicate with robots~\cite{muhammad2019creating, hoang2021virtual, al2012furhat}. 
For example, the AR's trajectory will be updated to show the robot's intent when the a passerby approaches the robot~\cite{watanabe2015communicating}. 
Also, the shape-shifting wall can change the content on the robot based on the user's behavior and position~\cite{takashima2016study}.

%% file: 9-applications.tex
\ifdouble
\FigureApplication
\fi

\coloredSection{brown}{Application Domains}
We identified a range of different application domains in AR and robotics.  
\textcolor{color8}{Figure~\ref{fig:applications}} summarizes the each category and the list of related papers. 
We classified the existing works into the following high-level application-type clusters:
\textcolor{color8}{{\it 1) domestic and everyday use, 2) industry applications, 3) entertainment, 4) education and training, 5) social interaction, 6) design and creative tasks, 7) medical and health, 8) telepresence and remote collaboration, 9) mobility and transportation, 10) search and rescue, 11) robots for workspaces, and 12) data physicalization.}}

\ifsingle
\FigureApplication
\fi

Detailed lists of application use cases within each of these categories can be found in \textcolor{color8}{Figure~\ref{fig:applications}}, as well as appendix, including detailed lists of references we identified.
The largest category is \textit{industry}. 
For example, industry application includes manufacturing, assembly, maintenance, and factory automation. 
In many of these cases, AR can help the user to reduce the assembly or maintenance workload or program the robots for automation. 
Another large category we found emerging is \textit{domestic and everyday use scenarios}. For example, AR is used to program robots for household tasks. Also, there are some other sub-categories, such as photography, tour guide, advertisement, and wearable robots. 
\textit{Games and entertainment} are popular with robotic user interfaces.
In these examples, the combination of robots and AR is used to provide an immersive game experience, or used for storytelling, music, or museums. 
Figure~\ref{fig:applications} suggests that there are less explored application domains, which can be investigated in the future, which include design and creative tasks, remote collaboration, and workspace applications.

%% file: 10-evaluation.tex
\coloredSection{black}{Evaluation Strategies}
In this section, we report our analysis of evaluation strategies for augmented reality and robotics. The main categories we identified are following the classification by Ledo et al.~\cite{ledo2018evaluation}: \textcolor{color9}{1) evaluation through demonstration, (2) technical evaluations, and (3) user evaluations}. The goal of this section is to help researchers finding the best technique to evaluate their systems, when designing AR for robotic systems. 

\sub{Evaluation-1. Evaluation through Demonstration}
\textit{Evaluation through demonstration} is a technique to see how well a system will potentially work in certain situations. The most common approach from our findings include showing example applications~\cite{yamaoka2016mirageprinter, scheible2013displaydrone, chou2004augmented, leithinger2011direct, garcia2011educational, kasahara2013extouch, renner2018facilitating, suzuki2016gushed, follmer2013inform,  ostanin2018interactive} and proof-of-concept demonstrations of a system~\cite{jones2021ar,araiza2019augmented,everitt20193d,young2006mixed,kojima2006augmented,carroll2013augmented,luebbers2019augmented,guhl2017concept,muhammad2019creating,rozenberszki2021towards}. Other common approaches include demonstrating a system through a workshop~\cite{yamaoka2016mirageprinter, leitner2010physical, leithinger2010relief}, demonstrating the idea to a focus group~\cite{xiao2013mirrorfugue, zhu2016virtually, renner2018wysiwicd, abbas2012augmented}, carrying out case studies~\cite{dinh2017augmented, hwang2008augmented, ong2006methodologies}, and providing a conceptual idea~\cite{xiao2013mirrorfugue, zhu2016virtually, renner2018wysiwicd, jost2018safe, roudaut2013morphees}.  


\sub{Evaluation-2. Technical Evaluation}
\textit{Technical Evaluation} refers to how well a system performs based on internal technical measures of the system. The most common approaches for technical evaluation are measuring latency~\cite{cao2019ghostar, calandra2021evaluating, zolotas2018head, cha2018effects, chan2020towards, nunez2006human}, accuracy of tracking~\cite{cha2018effects, cauchard2019drone, aleotti2017detection, chadalavada2020bi, zhang2020vision}, and success rate~\cite{liu2018interactive, nilwong2020outdoor}. Also, we found some works evaluate their system performances based on the comparison with other systems, which for example, include comparing tracking algorithms with other approaches~\cite{cha2018effects, bentz2019unsupervised, frank2016realizing, st2015robot, chacko2019augmented}.

\sub{Evaluation-3. User Evaluation}
\textit{User evaluation} refers to measuring the effectiveness of a system through user studies. To measure the user performance when interacting with the system, there are many different approaches and methods that are used. For example, the NASA TLX questionnaire is a very popular technique for user evaluation~\cite{chan2020towards, aleotti2017detection, quintero2018robot, chan2020augmented, chacko2019augmented}, which can be found used mostly for industry related applications. Other approaches include running quantitative~\cite{bentz2019unsupervised, huang2019flight, hedayati2018improving} and qualitative~\cite{andersen2016projecting, hamilton2021s, ganesan2018better} lab studies, through interviews~\cite{chadalavada2020bi,  diehl2020augmented, urbani2018exploring} and questionnaires~\cite{calandra2021evaluating, zolotas2018head, cha2018effects}. 
Sometimes systems combine user evaluations techniques with demonstration~\cite{scheible2013displaydrone, el2021teaching} or technical evaluations~\cite{aleotti2017detection, st2015robot}. 
In observational studies~\cite{cauchard2019drone, scheible2013displaydrone, robert2012blended}, researchers can also get user feedback through observations~\cite{von2016robot, fung2011augmented}. Finally, some systems also ran lab studies through expert interviews~\cite{pedersen2011tangible, ahn2013supporting, everitt2017polysurface} to get specific feedback from the expert's perspectives.

%% file: 11-discussion.tex
\section{Discussion and Findings}

\begin{figure*}[h!]
\centering
\includegraphics[width=\textwidth]{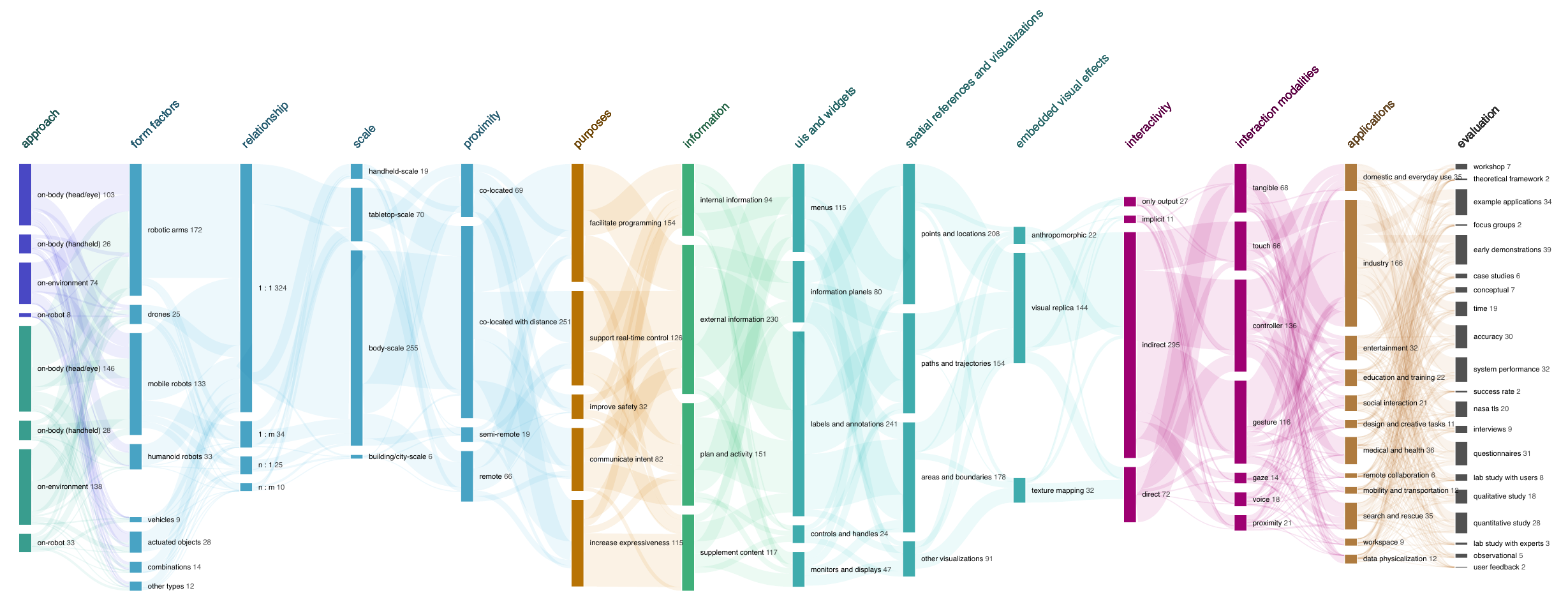}
\caption{A visualization with overall counts of characteristics across all dimensions.}%
\label{fig:summary}%
\end{figure*}


Based on the analysis of our taxonomy, Figure~\ref{fig:summary} shows a summary of the number of papers for each dimension.
In this section, we discuss common strategies and gaps across characteristics of selected dimensions.



\sub{Robot - Proximity Category} In terms of proximity, \textit{co-located with distance} are the preferred method in AR-HRI systems (251 papers). This means that the current trend
for AR-HRI systems is to have users co-located with the robot, but to not make any sort of contact with it. This also suggests that AR devices provide a promising way to interact with robots without having the need to directly make contact with it, such as performing robotic manipulation programming through AR~\cite{ostanin2020human}.

\sub{Design - UI and Widget Category}
In terms of the \textit{Design - UI and Widgets} category, labels and annotations are the most common choice (241 papers) used in AR systems. Given that AR enables us to design visual feedback without many of the constraints
of physical reality, researchers of AR-HRI systems take advantage of that fact to provide relevant information about the robot and/or other points of interest such as the environment and objects~\cite{darbar2019dronesar}. Increasingly, other widgets are used, such as information panels, floating displays, or menus. It is notable that only 24 papers made use of virtual control handles, possibly implying that AR is not yet commonly used for providing direct control to robots. 

\sub{Interactions - Level of Interactivity Category} 
For the \textit{Interaction Level} category, we observed that \textit{explicit and indirect input} is the most common approach within AR-HRI systems (295 papers). This means that user input through AR to interact with the robot must go through some sort of input mapping to accurately interact with the robot. This is an area that should be further explored, which we mention in \textbf{Section 11 - Immersive Authoring and Prototyping Environments for AR-HRI}. However, while AR may not be the popular approach in terms of controlling a robot's movement, as mentioned above, it is still an effective medium to provide other sorts of input, such as path trajectories~\cite{quintero2018robot}, for robots.

\sub{Interactions - Modality Category} 
In the \textit{Interaction Modality} category, \textit{pointers and controllers} (136 papers) and \textit{spatial gestures} (116 papers) are most commonly used. Spatial gestures, for example, are used in applications such as robot gaming~\cite{lupetti2018design}. Furthermore, \textit{touch} (66 papers) and \textit{tangibles} (68 papers) are also common interaction modalities, indicating that these traditional forms of modality are seen as effective options for AR-HRI systems (for example, in applications such as medical robots~\cite{wen2017augmented} and collaborative robots~\cite{vzidek2021cnn}). It is promising to see how many AR-HRI systems are using tangible modalities to provide shape-changing elements~\cite{leithinger2013sublimate} and control~\cite{pedersen2011tangible} to robots. Gaze and voice input are less common across the papers in our corpus, similar to proximity-based input, pointing to interesting opportunities for future work to explore these modalities in the AR-HRI context.

%% file: 12-opportunity.tex
\section{Future Opportunities}

Finally, we formulate open research questions, challenges, and opportunities for AR and robotics research.
For each opportunity, we also discuss potential research directions, providing sketches and relevant sections or references as a source of inspiration. We hope this section will guide, inspire, and stimulate the future of AR-enhanced Human-Robot Interaction (AR-HRI) research.

\FigureOpportunityOne

\opportunity{1}{Making AR-HRI Practical and Ubiquitous}

\opportunitysub{Technological and Practical Challenges}
While AR-HRI has a great promise, there are many technological and practical challenges ahead of us. 
For example, the accurate realistic superimposition or occlusion of virtual elements is still very challenging due to noisy real-time tracking.
The \textbf{\textit{improvement of display and tracking technologies}} would broaden the range of practical applications, especially when more precise alignments are needed, such as robotic-assisted surgery or medical applications (\textbf{Section 9.7}). 
Moreover, \textbf{\textit{error-reliable system design}} is also important for practical applications.
AR-HRI is used to improve safety for human co-workers (\textbf{Section 5.2}), however, if the AR system fails in such a safety-critical situation, users might be at risk (e.g., device malfunctions, content misalignment, obscured critical objects with inappropriate content overlap, etc).
It is important to \textbf{\textit{increase the reliability}} of AR systems from both systems design and user interaction perspectives (e.g., What extent should users rely on AR systems in case the system fails? How can we avoid visual clutter or the occlusion of critical information in a dangerous area? etc).
These technical and practical challenges should be addressed before we can see AR devices be common in everyday life.

\opportunitysub{Deployment and Evaluation In-the-Wild}
Related to the above, most prior AR-HRI research has been done in controlled laboratory conditions. 
It is still questionable whether these systems and findings can be directly applied to a real-world situation.
For example, outdoor scenarios like search-and-rescue or building construction (\textbf{Section 9}) may require very different technical requirements than indoor scenarios (e.g., Is projection mapping visible enough outdoors? Can outside-in tracking sufficiently cover the area that needs to be tracked?).
On the other hand, the current HMD devices still have many usability and technical limitations, such as display resolution, visual comfort, battery life, weight, the field of view, and latency issues.
To appropriately design a practical system for real-world applications, it is important to design based on the user's needs through \textbf{\textit{user-centered design}} by conducting a repeated cycle of interviews, prototyping, and evaluation.
In particular, researchers need to carefully consider different approaches or technological choices (\textbf{Section 3}) to meet the user's needs. 
The deployment and evaluation in the wild will allow us to develop a better understanding of what kind of designs or techniques should work and what should not in real-world situations.


\FigureOpportunityTwo

\opportunity{2}{Designing and Exploring New AR-HRI}

\opportunitysub{Re-imagining Robot Design without Physical Constraints}
With AR-HRI, we have a unique opportunity to \textbf{\textit{re-imagine robots design}} without constraints of physical reality.
For example, we have covered interesting attempts from the prior works, like \textit{making non-humanoid robots humanoid}~\cite{hiroi2010evaluation, jones2021belonging, young2007robot} or \textit{making robots visually animated}~\cite{hololens-robot, groechel2019using, al2012furhat} (\textbf{Section 7.3}), either through HMD~\cite{jones2020vroom} or projecion~\cite{al2012furhat} (\textbf{Section 3}). 
However, this is just the tip of the iceberg of such possibilities.
For example, what if robots would look like a fictional character~\cite{cauchard2021drones, kari2021transformr} or behave like Disney's character animation?~\cite{kazi2016motion, thomas1995illusion, van2004bringing}
We believe there is still a huge untapped design opportunity for \textbf{\textit{augmented virtual skins}} of the robots by fully leveraging the unlimited visual expressions.
In addition, there is also a rich design space of \textbf{\textit{dynamic appearance change}} by leveraging visual illusion~\cite{lindlbauer2017changing, lindlbauer2018remixed}, such as making robots disappear~\cite{morin2012camouflage, robert2012blended}, change color~\cite{wang2016mechanical, gossweiler2015mechanochemically}, or transform its shape~\cite{suzuki2019shapebots, hedayati2020pufferbot, suzuki2021hapticbots, shah2021shape} with the power of AR.
By increasing the expressiveness of robots (\textbf{Section 5.5}), this could improve the engagement of the users and enable interesting applications (e.g., using drones that have facial expression~\cite{herdel2021drone} or human body/face~\cite{gomes2016bitdrones} for remote telepresence~\cite{jones2020vroom}).
We argue that there are still many opportunities for such \textbf{\textit{unconventional robot design}} with expressive visual augmentation.
We invite and encourage researchers to re-imagine such possibilities for the upcoming AR/MR era.

\opportunitysub{Immersive Authoring and Prototyping Environments for AR-HRI}
Prototyping functional AR-HRI systems is still very hard, given the high barrier of requirements in both software and hardware skills.
Moreover, the development of such systems is pretty time-consuming---people need to continuously move back and forth between the computer screen and the real world, which hinders the rapid design exploration and evaluation.
To address this, we need a better authoring and prototyping tool that allows \textbf{\textit{even non-programmers}} to design and prototype to broaden the AR-HRI research community.
For example, what if, users can design and prototype interactions \textbf{\textit{through direct manipulation within AR}}, rather than coding on a computer screen? (e.g., one metaphor is, for example, Figma for app development or Adobe Character Animator for animation)
In such tools, users also must be able to design \textbf{\textit{without the need for low-level robot programming}}, such as actuation control, sensor access, and networking.
Such AR authoring tools have been explored in the HCI context~\cite{billinghurst2021rapid, leiva2020pronto, nebeling2018protoar, wang2020capturar} but still relatively unexplored in the domain of AR-HRI except for a few examples~\cite{cao2019ghostar, suzuki2018reactile}.
We envision the future of intuitive authoring tools should invoke further design explorations of AR-HRI systems (\textbf{Section 7}) by democratizing the opportunity to the broader community. 


\FigureOpportunityThree

\opportunity{3}{AR-HRI for Better Decision-Making}

\opportunitysub{Real-time Embedded Data Visualization for AR-HRI}
AR interfaces promise to support operators' complex decision-making (\textbf{Section 5.2}) by aggregating and visualizing various data sources, such as internal, external, or goal-related information (\textbf{Section 6.1-6.3}).
Currently, such visualizations are mainly limited with simple spatial references of user-defined data points (\textbf{Section 7.2}), but there is still a huge potential to connect data visualization to HRI~\cite{szafir2021connecting} in the context of AR-HRI. 
For example, what if AR interfaces can \textbf{\textit{directly embed real-time data onto the real world}}, rather than on a computer screen?
We could even combine real-time visualizations with a \textbf{\textit{world-in-miniature}}~\cite{danyluk2021design} to facilitate navigation in a large area, such as drone navigation for search-and-rescue.
We can take inspiration from existing immersive data analysis~\cite{chandler2015immersive, ens2021grand} or real-time embedded data visualization research~\cite{willett2016embedded, suzuki2020realitysketch, willett2021perception} to better design such data-centric interfaces for AR-HRI. 
We encourage the researchers to start thinking about how we can apply these emerging data visualization practices for AR-HRI systems in the future.

\opportunitysub{Explainable and Explorable Robotics through AR-HRI}
As robots become more and more intelligent and autonomous, it becomes more important to make the robot's decision-making process visible and interpretable.
This is often called \textit{Explainable AI} in the context of machine learning and AI research, but it is also becoming relevant to robotics research as \textbf{\textit{Explainable Robotics}}~\cite{setchi2020explainable, das2021explainable}.
Currently, such explanations are represented as descriptive text or visuals on a screen~\cite{das2021explainable}.
However, by leveraging AR-HRI systems, \textbf{\textit{users can better understand the robots' behavior}} by seeing what they see (sensing), how they think (decision making), and how they respond (actions) in the real world.
For example, what if users can see what a robot recognizes as obstacles or how it chooses the optimal path when navigating in a crowded place?
More importantly, these visualizations are also \textbf{\textit{explorable}}---users can interactively explore to see how the robot's decision would change when the physical world changes (e.g., directly manipulating physical obstacles to see how the robot's optimal path updates). Such interfaces could help programmers, operators, or co-workers understand the robot's behavior more easily and interactively.
Future research should \textbf{\textit{connect explainable robotics with AR}} to better visualize the robot's decision-making process \textit{embedded in real-world}. 


\FigureOpportunityFour


\opportunity{4}{Novel Interaction Design enabled by AR-HRI}

\opportunitysub{Natural Input Interactions with AR-HRI Devices}
With the proliferation of HMD devices, it is now possible to use expressive inputs more casually and ubiquitously, including gesture, gaze, head, voice, and proximity-based interaction (\textbf{Section 8.2}).
In contrast to environment-installed tracking, HMD-based hand- and gaze-tracking could enable more natural interactions without the constraint of location.
For example, with the hand-tracking capability, we can now implement expressive gesture interactions, such as finger-snap, hand-waving, hand-pointing, and mid-air drawing for swarm drone controls in entertainment, search and rescue, firefighting, or agricultural foraging~\cite{alon2021drones, kim2020user}).
In addition, the combination of multiple modalities, such as voice, gaze, and gesture is also an interesting direction. 
For example, when the user says \textit{``Can you bring this to there?''}, it is usually difficult to clarify the ambiguity (e.g., ``this'' or ``there''), but with the combination of gaze and gesture, it is much easier to clarify these ambiguities within the context.
AR-based visual feedback could also help the user clarify their intentions. The user could even casually register or program such a new input on-demand through end-user robot programming (\textbf{Section 5.1}).
Exploring new interactions enabled by AR-HRI systems is also an exciting opportunity. 

\opportunitysub{Further Blending the Virtual and Physical Worlds} 
As robots weave themselves into the fabric of our everyday environment, the term \textit{``robots''} no longer refer to only traditional humanoid or industry robots, but can become a variety of forms (\textbf{Section 2.1} and \textbf{Section 4.1})---from self-driving cars~\cite{jaguar-rover} to robotic furniture~\cite{suzuki2020roomshift, yixian2020zoomwalls}, wearable robots~\cite{dementyev2016rovables}, haptic devices~\cite{vonach2017vrrobot}, shape-changing displays~\cite{follmer2013inform}, and actuated interfaces~\cite{pangaro2002actuated}. 
These ubiquitous robots will be used to  \textbf{\textit{actuate our physical world}} to make the world more dynamic and reconfigurable. 
By levering both AR and this physical reconfigurability, we envision further blending virtual and physical worlds with a \textbf{\textit{seamless coupling between pixels and atoms}}.
Currently, AR is only used to \textit{visually augment} appearances of the physical world. However, what if \textbf{\textit{AR can also ``physically'' affect the real-world}}? For example, what if a \textit{virtual user} pushes a physical wall then it moves synchronously? What if \textit{virtual wind} can wave a physical cloth or flag? What if \textit{virtual explosion} can make a shock wave collapse physical boxes? 
Such \textbf{\textit{virtual-physical interactions}} would make AR more immersive with the power of visual illusion, which can also have some practical applications such as entertainment, remote collaboration, and education. 
Previously, such ideas were only partially explored~\cite{aoki2005kobito, siu2018investigating}, but we believe there still remains a rich design space to be further explored.
For future work, we should further seek to blend virtual and physical worlds by leveraging both visually (AR) and physically (robotic reconfiguration) programmable environments.

%% file: 13-conclusion.tex
\section{Conclusion}
In this paper, we present our survey results and taxonomy of AR and robotics, synthesizing existing research approaches and designs in the eight design space dimensions.
Our goal is to provide a common ground for researchers to investigate the existing approaches and design of AR-HRI systems.
In addition, to further stimulate the future of AR-HRI research, we discuss future research opportunities by pointing out eight possible directions: 1) technological and practical challenges, 2) deployment and evaluation in-the-wild, 3) re-imagining robot design, 4) immersive authoring and prototyping environments, 5) real-time embedded data visualization for AR-HRI, 6) explainable and explorable robotics with AR, 7) novel interactions techniques, and 8) further blending the virtual and physical worlds with programmable augmentation and actuation. 
We hope our survey, taxonomy, and open research opportunity will guide and inspire the future of AR and robotics research.

%% file: 14-appendix.tex

\begin{table*}[t]
\centering
\caption*{Appendix Table: Full Citation List}
\label{tab:full-list}
\TableFontSize
\begin{tabular}{ \TableConfig }
\TableHeader \\
\hline
\\
\multicolumn{3}{l}{
{\bf Section 3. Approaches to Augmenting Reality in Robotics}
} \\
\\
\textcolor{color1}{
\textbf{\textit{Augment Robots}}
}
& &
\\
\end{tabular}
\end{table*}

\begin{table*}[t]
\centering
\TableFontSize
\begin{tabular}{ \TableConfig }
\TableHeader \\
\hline
\\
\multicolumn{3}{l}{
{\bf Section 4. Characteristics of Augmented Robots}
} \\
\\
\textcolor{color3}{
\textbf{\textit{Form Factors}}
}
& &
\\
\end{tabular}
\end{table*}

\begin{table*}[t]
\centering
\TableFontSize
\begin{tabular}{ \TableConfig }
\TableHeader \\
\hline
\\
\multicolumn{3}{l}{
{\bf Section 5. Purposes and Benefits of Visual Augmentation} 
} \\
\\
\end{tabular}
\end{table*}

\begin{table*}[t]
\centering
\TableFontSize
\begin{tabular}{ \TableConfig }
\TableHeader \\
\hline
\\
\multicolumn{3}{l}{
{\bf Section 7. Design Components and Strategies} 
} \\
\\
\textcolor{color6}{
\textbf{\textit{UIs and Widgets}}
}
& &
\\
\end{tabular}
\end{table*}

\begin{table*}[t]
\centering
\TableFontSize
\begin{tabular}{ \TableConfig }
\TableHeader \\
\hline
\\
\multicolumn{3}{l}{
{\bf Section 8. Interactions} 
} \\
\\
\textcolor{color7}{
\textbf{\textit{Interactivity}}
}
& &
\\
\textcolor{color7}{
$\rightarrow$ Only Output
}
& \textcolor{color7}{
27 
} &
Figure: \cite{walker2018communicating}
---
\cite{walker2018communicating, young2007robot, groechel2019using, suzuki2016gushed, tobita2011floating, scheible2013displaydrone, rosen2020communicating, lingamaneni2017dronecast, papachristos2016augmented, makhataeva2019safety, topliss2018establishing, nilwong2020outdoor, fujii2020development, hiroi2010evaluation, williams2019mixed, caruso2010robotic, lee2020realization, dragone2006mixing, chadalavada2020bi, jones2020vroom, reardon2019communicating, corotan2019indoor, korthauer2020watch, everitt20193d, ghiringhelli2014interactive, tavares2019collaborative, bianchi2021use}
\\
\textcolor{color7}{
$\rightarrow$ Implicit
}
& \textcolor{color7}{
11 
} &
Figure: \cite{watanabe2015communicating}
---
\cite{bolano2019transparent, watanabe2015communicating, lindlbauer2016combining, subin2017android, qiu2020human, bentz2019unsupervised, strazdas2021robo, vzidek2021cnn, michalos2016augmented, coovert2014spatial, lera2014augmented}
\\
\end{tabular}
\end{table*}

\begin{table*}[t]
\centering
\TableFontSize
\begin{tabular}{ \TableConfig }
\TableHeader \\
\hline
\\
\multicolumn{3}{l}{
{\bf Section 9. Application Domains} 
} \\
\\
\textcolor{color8}{
Domestic and Everyday Use
}
& \textcolor{color8}{35
} &
Figure: \cite{liu2011roboshop}
---
{\it Household Task}
authoring home automation~\cite{el2021teaching, fung2011augmented, lakshantha2014human, seiger2017mixed, ishii2009designing, hiraki2019navigatorch, kato2009multi, cao2019v, liu2011roboshop, guo2009touch, ostanin2019interactive, ren2019augmented, young2011mixed}, item movement and delivery~\cite{liu2020mobile, dragone2007using, kasahara2013extouch, chen2021pinpointfly}, displaying cartoon-art while sweeping~\cite{young2007robot}, multi-purpose table~\cite{takashima2013transformtable}
{\it Photography}
drone photography~\cite{hedayati2018improving,erat2018drone},
{\it Advertisement}
mid-air advertisement~\cite{nozaki2014flying, scheible2013displaydrone, suzuki2016gushed, yamada2017isphere} 
{\it Wearable Interactive Devices}
haptic interaction~\cite{urbani2018exploring}, fog screens~\cite{suzuki2016gushed}, head-worn projector for sharable AR scenes~\cite{hartmann2020aar}
{\it Assistance and companionship}
elder care~\cite{chae2018pervasive}, personal assistant~\cite{lee2011note, ro2019projection, renner2018wysiwicd}
{\it Tour and exhibition guide}
tour guide~\cite{cauchard2019drone, mohareri2011autonomous}, museum exhibition guide~\cite{ro2019projection, hartmann2020aar}, guiding crowds~\cite{yamada2017isphere}, indoor building guide~\cite{corotan2019indoor}, museum interactive display~\cite{elsharkawy2021uwb}
\\
\end{tabular}
\end{table*}